\documentclass[11pt,a4paper]{article}
\usepackage[hyperref]{emnlp-ijcnlp-2019}
\usepackage{times}
\usepackage{latexsym}
\usepackage{scrextend}

\usepackage{amsmath}

\usepackage{url}

\usepackage{tikz}
\usetikzlibrary{shapes,arrows}
\tikzstyle{decision} = [diamond, draw, fill=blue!20, 
    text width=4.5em, text badly centered, node distance=3cm, inner sep=0pt]
\tikzstyle{block} = [fill=blue!30, inner sep=7pt, text centered]
\tikzstyle{line} = [draw, -latex']
\tikzstyle{cloud} = [draw, ellipse,fill=red!20, node distance=3cm, 
minimum height=2em]
\usetikzlibrary{positioning, fit}
\usetikzlibrary{shapes,snakes}

\usepackage{caption}
\usepackage{subcaption}
\aclfinalcopy % Uncomment this line for the final submission

\title{Learning to Request Guidance in Emergent Communication}

\author{Benjamin Kolb\footnotemark[1], Leon Lang\thanks{ \ \ Equal contributions}, Henning Bartsch, Arwin Gansekoele, \\
\textbf{Raymond Koopmanschap, Leonardo Romor, David Speck, Mathijs Mul\thanks{ \ \ Shared senior authorship}, Elia Bruni\footnotemark[2]} \\
\\
\tt \{benjamin.kolb, leon.lang, henning.bartsch, \\
\tt arwin.gansekoele, raymond.koopmanschap,  leonardo.romor, \\
\tt  david.speck, mathijs.mul\}@student.uva.nl
\\
{\tt elia.bruni@gmail.com} \\
\\
University of Amsterdam
}

\date{}

\begin{document}

\maketitle

\begin{abstract}

%Prior work has shown that the learning process of an Imitation Learning learner can successfully be sped up by messages received from a pretrained guide network. Since guidance is costly in the real world we allow the learner to make a binary decision whether it wishes to receive guidance and give it a small penalty for guidance requests.
%We find that the learner first uses the guidance to speed up its learning process and later reduces the frequency of the guidance requests significantly. We furthermore investigate the situations where guidance is requested and find that those are indeed those where the guidance is most valuable.

%Previous research into agent communication has shown that a pre-trained guide can speed up the learning process of an imitation learning agent. The guide achieves this by providing the agent with discrete messages in an emerged language about how to solve the task. We extend this research by a one-bit communication channel from the learner back to the guide: It is able to ask the guide for help, and we limit the guidance by penalizing the learner for these requests. We show that over time the learner manages to ask for help in those situations where additional information is especially worthwhile. This research is a first step into analyzing the dynamics that emerge from more complicated communication protocols between different entities, and over the long run might help shedding light on communication protocols involving humans.

Previous research into agent communication has shown that a pre-trained guide can speed up the learning process of an imitation learning agent. The guide achieves this by providing the agent with discrete messages in an emerged language about how to solve the task. We extend this one-directional communication by a one-bit communication channel from the learner back to the guide: It is able to ask the guide for help, and we limit the guidance by penalizing the learner for these requests. During training, the agent learns to control this gate based on its current observation. We find that the amount of requested guidance decreases over time and guidance is requested in situations of high uncertainty. We investigate the agent's performance in cases of open and closed gates and discuss potential motives for the observed gating behavior.

%Prior work showed that a pretrained guide agent can speed up the learning process of an imitation learning agent. We extend this one-directional communication between guide and learner by giving the learner the possibility to produce a binary signal. With this signal, it can indicate in every situation whether it needs guidance or not. We incentivize the learner with a penalty to only ask for guidance when the added information is valuable. During training, the agent learns to control this gate based on its current observation. We find that the amount of requested guidance decreases over time and guidance is requested in situations of high uncertainty. We investigate the agent's performance in cases of open and closed gates and discuss potential motives for the observed gating behavior.
\end{abstract}

\section{Introduction}\label{Introduction}

% A long-term goal of AI is to create agents that are able to autonomously solve complex tasks in the real world. 
A long-term goal of AI is to develop agents that can help humans execute complex tasks in the real world. Since reward functions that are aligned with human intentions are hard to manually specify \citep{faulty}, other approaches besides Reinforcement Learning are needed for creating agents that behave in the intended way. 
% a useful way. 
Among these are Reward Modeling \citep{reward_modeling} and Imitation Learning \citep{imitation_learning}, but, eventually, it would be useful if we could use natural language to transmit wishes to the agents. 

Recently, \citet{babyguide} made progress in this direction by showing how communication can be used to guide a learner in a gridworld environment. Using emergent discrete messages, the guide is able to speed up the learning process of the learner and to let it generalize across incrementally more difficult environments. 

% In order to simulate agent communication in a simple setting, \citet{babyguide} uses a guide trained to transmit simple messages from an emergent discrete language in order to help a learner achieve its goal in gridworld environments from \citet{babyai}. Among other things, they show that guidance helps to speed up the imitation learning process.
In this setting, the communication channel is completely one-way: in each time step, the guide transmits a message that may help the learner make its decisions. In reality, however, communication is more complex than that: the guidance may be expensive and it can, therefore, be beneficial to have more sparse messages. Furthermore, the learner may want to ask for clarification if something is unclear. Therefore, it would be worthwhile if there was a communication channel from the learner back to the guide. It is this interactive nature of communication that arguably is needed for more advanced AI systems \citep{mikolov}.

In this paper, we equip the learner introduced by \citet{babyguide} with a binary gate to indicate its need for guidance in each time step. A penalty for the use of guidance incentivizes a sparse usage of the gate. By analyzing the relationship between the learner's usage of the gate and a number of measures, we show that the learner indeed learns to ask for guidance in a smart and economical way.

\section{Related Work}
\label{sec:related}
%Think about:
%\begin{itemize}
%    \item agent communication
%    \item following instructions
%    \item emergent language
%    \item imitation learning
%    \item guidance request / two-way communication / acting different when uncertain
%\end{itemize}{}
In this section we briefly lay out relevant work that relate to our approach on the dimensions of following language instructions, emergent communication and the interactions that emerge from guidance requests.
%Try to reduce Mathijs citations by 1/3 and look for new (and relevant) papers. Aim for a section length of roughly 1 to max 1.5 columns

\subsection{Following Language Instruction}
In recent years, much research has been conducted in the field of following language instructions. Since language commands build a way to interact with agents and communicate information in an effective and human-interpretable way, the agent's processing and understanding of these commands is relevant for our project. Starting with manually engineered mappings \citep{winograd1971procedures}, the currently most relevant grounded language acquisition methods focus on learning a parser to map linguistic input to its executable equivalent, i.e. action specifications using statistical models \citep{yu2018interactive}, \citep{kim2012unsupervised}, \citep{artzi2013weakly}, \citep{mei2016listen}.
In the BabyAI platform introduced by \citet{babyai} a synthetic ``Baby Language'' is used, which consists of a subset of English and whose semantics is generated by a context-free grammar (as opposed to instruction templates) and is easily understood by humans. They employ a single model to combine linguistic and visual information, similar to \citet{misra2017mapping}. Our setup builds on \citet{babyguide} who extend that platform with a guide, like \citet{co2018guiding}, that supplements the agent's information with iterative linguistic messages.

\subsection{Emergent Communication}
In order to benefit most from the guide, the agent would ideally communicate back, thus creating a multi-agent cooperation scenario. Recent research in this area investigates the emergence and usage of emergent language, e.g. in the context of referential games \citep{lazaridou2016multi}. Furthermore, \citet{mordatch2018emergence} show that multiple agents can develop a grounded compositional language to fulfill their tasks more effectively with spoken exchange. In our setup the emergent communication consists of discrete word tokens similar to \citet{havrylov2017emergence}. \citet{jiang2018learning} propose an attentional communication model to learn when communication is needed (and helpful), resulting in more effective (large-scale) multi-agent cooperation. 

\subsection{Guidance Requests}
In prior work \citep{babyguide}, guidance is given at every time step and the communication is one-way from guide to learner. We extend this approach by allowing a communication channel in the other direction. Here we survey work that uses similar requests for help.

Most similar to our work is \citet{1997}, where ``Ask for help'' is proposed: in this setting, an RL agent has one additional action with which it can signify to a ``trainer'' that it wants to receive help. The trainer then chooses the agent's action. Whether to ask for help is based on uncertainty about the highest action value. This is different from our setting in which the uncertainty is only \emph{implicitly} responsible for queries, as can be seen in Section \ref{sec:Analysis}. \citet{kosoy} studies the ``Ask for help'' setting theoretically and proves a regret bound for agents that act in infinite-horizon discounted MDPs and are able to delegate actions to an ``advisor''.

In \citet{nguyen2019vision} there is a help-requesting policy $\pi_{\text{help}}$ that can signify if the agent needs help. If this is the case, a guide answers with a language-based instruction of subgoals. Additionally, there is a budget that limits asking for help.

Also related is \citet{on_the_job}, where structured prediction problems are considered: a sequence of words is received and each word is supposed to be mapped to a label. The system can query a crowd (as in crowd-sourcing) to obtain answers on specific words in the sequence. As in our case, querying the crowd is penalized by an additional loss.

In \citet{active_reinforcement_learning}, \citet{arl2}, active reinforcement learning (ARL) is proposed: different from usual RL, the agent has to choose in each round if it \emph{wants} to receive the reward that results from its action, which results in a constant query-cost $c > 0$. Note that in this setting, what is queried is feedback, whereas in our setting, the model queries guidance \emph{prior} to making a decision. Active Reward Learning \citep{active_reward_learning} is a similar approach in the context of continuous control tasks.

\section{Approach}
\subsection{BabyAI Game}
% About imitation learning: Basically as in Mathijs paper. Note that it may be confusing for the reader that we have an expert that gives data and an additional guide that gives input (and as a third layer a language instruction!). Super confusing for anyone not in the area.

All our experiments take place in the BabyAI platform \citep{babyai}. In this platform, an agent learns to complete tasks given by instructions in a subset of English in a mini gridworld environment. The environments are only partially observable to the agent.
%Besides the possibility to train a learning agent on its own by reinforcement learning, it can also be trained by imitation learning (behavioral cloning) on expert data \citep{imitation_learning}, which is the approach we take in our work. In this setting, the learner is placed in all the situations that an expert faced and is judged by the cross-entropy loss between its distribution over actions and the ``true'' action by this expert. This effectively reduces the training process to supervised learning.

\begin{figure}
\centering
\begin{minipage}[b]{0.44\linewidth}
    \centering
    \includegraphics[width=\linewidth]{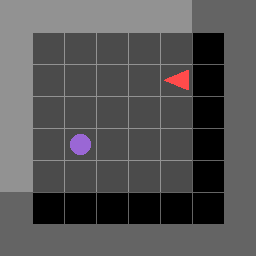}
    {(a) GoToObj} \label{fig:gotoobj}
\end{minipage}%
\hspace{20pt}
\begin{minipage}[b]{0.44\linewidth}
    \centering
    \includegraphics[width=\linewidth]{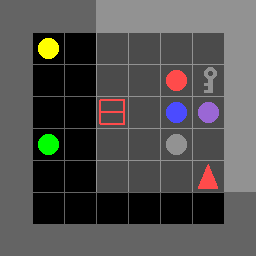}
    {(b) PutNextLocal} \label{fig:bosslevel}
\end{minipage}
\caption{Example levels from BabyAI. GoToObj is the simplest level, requiring the agent to go to a specific object. This task is given by a language instruction to the agent, such as ``Go to the purple ball''. PutNextLocal is more difficult, requiring a more complex skill-set. An example task in this setting would be ``Put the grey key next to the red box". Shown in the images are the agent (red triangle), different objects (boxes, keys, balls in different colors) and the observation of the agent (a brighter $7 \times 7$ visual field that has the agent in the middle of the bottom row relative to the agent. Parts of this is outside of the shown images)}%
\label{fig:babyai}%
\end{figure}

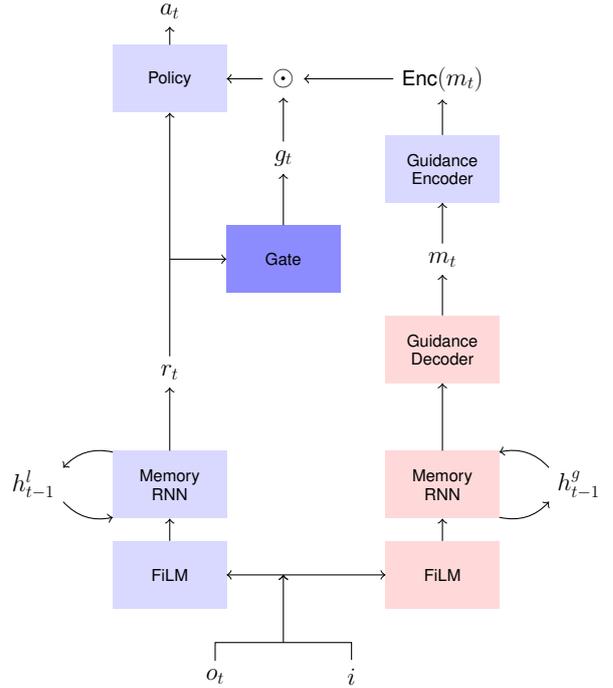
\begin{figure}[t]
\centering
\begin{center}
    \begin{tikzpicture}[scale=0.299, every node/.style={transform shape}]
% \begin{tikzpicture}[scale=0.32]

\tikzset{every node}=[font=\huge\sffamily]

% Inputs
\node[scale=1.5] at (-3,-1.5) (o) {$o_t$};
\node[scale=1.5] at (3,-1.5) (i) {$i$};
\node[scale=1.5] at (-5,28) (a) {$a_t$};
\node[scale=1.5] at (7,17) (m) {$m_t$};
\node[scale=1.5] at (-5,12) (r) {$r_t$};
\node[scale=1.5] at (0,21.5) (g) {$g_t$};
\node[scale=1.5] at (7, 25) (me) {$\text{\LARGE{Enc}}(m_t)$};

% memories nodes
\node[scale=1.5] at (-11,7) (meml) {$h^{l}_{t-1}$};
\node[scale=1.5] at (13,7) (memg) {$h^{g}_{t-1}$};

% dot
\node[scale=1.7] at (0, 25) (dot) {$\odot$};

% Learner blocks
\node at (-5,3) [rectangle,minimum height=3cm,minimum width=5cm,fill=blue!15] (flm){FiLM};
\node at (-5,7) [rectangle,minimum height=3cm,minimum width=5cm,fill=blue!15,align=center] (mrnn){Memory \\ RNN};
\node at (-5,25) [rectangle,minimum height=3cm,minimum width=5cm,fill=blue!15] (plcy){Policy};
\node at (0,17) [rectangle,minimum height=3cm,minimum width=5cm,fill=blue!45] (gate){Gate};
\node at (7,21) [rectangle,minimum height=3cm,minimum width=5cm,fill=blue!15,align=center] (genc){Guidance \\ Encoder};

% Guide blocks

\node at (7,3) [rectangle,minimum height=3cm,minimum width=5cm,fill=red!15] (flmg){FiLM};
\node at (7,7) [rectangle,minimum height=3cm,minimum width=5cm,fill=red!15,align=center] (mrnng){Memory \\ RNN};
\node at (7,13) [rectangle,minimum height=3cm,minimum width=5cm,fill=red!15,align=center] (gdec){Guidance \\ Decoder};

\draw[->] (flmg) edge (mrnng) (mrnng) edge (gdec) (gdec) edge (m)(m) edge (genc);
\draw[->] (genc) edge (me) (me) edge (dot) (dot) edge (plcy);
\draw[->] (flm) edge (mrnn) (mrnn) edge (r);
\draw[->] (gate) edge (g);
\draw[->] (-5, 17) to (gate);
\draw[->] (-5, 17) to (plcy);
\draw[->] (g) edge (dot);
\draw[->] (plcy) edge (a);
\draw (r) -- (-5, 17);

%\draw (mrnn) -- (0, 7);
%\draw[->] (0, 7) edge (gate);
%\draw[->, to path={|- (\tikztotarget)}]
%  (h) edge (gate);   

\draw[->] (mrnn) edge[bend right] node [left] {} (meml);
\draw[->] (meml) edge[bend right] node [left] {} (mrnn);

\draw[->] (mrnng) edge[bend right] node [left] {} (memg);
\draw[->] (memg) edge[bend right] node [left] {} (mrnng);

\draw[->] (0, 3) edge (flm);
\draw[->] (0, 3) edge (flmg);
\draw (0, 3) -- (0, 0);
\draw (i) -- (3,0) --  (0, 0);
\draw (o) -- (-3,0) --  (0, 0);
\end{tikzpicture}   
\end{center}
\caption{Architecture of a learner that can ask for guidance. Depicted variables are $o_t$: observation input, $i$: linguistic instruction, $h_{t-1}^{l}$ and $h_{t-1}^{g}$: memory, $r_t$: learned representation, $m_t$: the discrete guidance message, $g_t$: the gating weight and $a_t$: the action chosen based on $r_t$ and possibly the encoded message $\text{Enc}(m_t)$. The red part (the guide) is pretrained and then finetuned, while the blue parts (conceptually belonging to the learner) are newly initialized at the beginning of the training.}
\label{fig:architecture}
\end{figure}

Figure \ref{fig:babyai} shows two example levels. In total there are 19 different levels that increase in difficulty and complexity of tasks. For each level, the BabyAI framework can randomly generate many missions that require roughly the same skillset and are provided with a similar language instruction. For the following investigation, we only focus on the levels ``GoToObj'' and ``PutNextLocal''. These are chosen to be simple but nevertheless require a representative set of skills. As such, our results can be understood as a proof of concept.

\subsection{Model}

In this section, we describe the model that we use for a learner that may ask for guidance. We refer to Figure \ref{fig:architecture} for a visual explanation.

Based on the observation $o_t$, the instruction $i$ and its own memory unit, the learner $L$ builds a representation $r_t$: 
\begin{equation}
r_t = L(o_t, i).
\end{equation} 

$o_t$ is a $7 \times 7 \times 3$ tensor describing the viewable environment (e.g. the brighter area in Figure \ref{fig:babyai}) and $i$ is a natural language instruction (e.g. "Go to the purple ball"). The representation unit $L$ uses a FiLM module \cite{film} followed by a memory RNN. 

First, consider a learner without guidance. In this setting it directly takes the representation $r_t$ and feeds it into the policy module $P$ that outputs the next action $a_t = P(r_t)$. For more details on this baseline setting, see \citet{babyai}.

\subsubsection*{A guided learner}

Now consider \citet{babyguide}, where a guide is added. The guide follows the same architecture as the learner-policy combination, but between the representing unit and the policy there is a discrete ``bottleneck'' that only allows $9$ different messages to pass. The policy then needs to encode this message continuously in order to choose the correct action out of $7$ possibilities. After this guide-policy combination is trained, the messages are fed into the policy attached to a newly initialized learner in order to help it make its decision. In this later guided training stage, the policy of the guide is not used anymore.

More formally, the guide uses the same input and a memory unit to produce a message $m_t$ of two words with 3 possible tokens for each.
\begin{equation}
m_t = G(o_t, i).
\end{equation}
The message $m_t$ is then encoded to a higher dimensional continuous encoding $\text{Enc}(m_t)$ that is produced by an encoder of the same architecture as the encoder used while training the guide. The policy then bases its decision on both the learned representation $r_t$ and encoding $\text{Enc}(m_t)$, which are simply concatenated:
\begin{equation}
a_t = P(r_t, \text{Enc}(m_t)).
\end{equation}
More details can be found in \citet{babyguide}.

\subsubsection*{Adding a gate}

%To enable the learner to decide when to receive guidance, we extend the learner with a gate module $G$ to learn a gating weight $g_t \in \{0, 1\}$ that switches the policy input between $(r_t, \text{Enc}(m_t))$ (guided) and $(r_t, \mathbf{0})$ (unguided):
To enable the learner to decide when to receive guidance, we extend the learner with a gate module $G$ to learn a gating weight $g_t \in \{0, 1\}$ that switches the policy input between $(r_t, \text{Enc}(m_t))$ (guided) and $(r_t, \mathbf{0})$ (unguided):
\begin{align}
\begin{split}
a_t & = P(r_t, g_t \cdot \text{Enc}(m_t)) \\
& = P\left((1 - g_t) \cdot (r_t, \mathbf{0}) + g_t \cdot (r_t, \text{Enc}(m_t))\right) \\
& = \begin{cases}
P(r_t, \mathbf{0}), & \text{if } g_t = 0, \\
P(r_t, \text{Enc}(m_t)), & \text{if } g_t = 1,
\end{cases}
\end{split}
\end{align}

%The gating weight $g_t$ is the result of a two-layered MLP with tanh activations applied to $r_t$, followed by another sigmoid activation and a binary comparison whether the result is $\geq 0.5$. The module that produces the gating weight will from here on be referred to as the gate. 
where $g_t = G(r_t)$. The gate module $G$ is a two-layered MLP with a tanh activation functions that outputs a scalar, followed by a sigmoid activation function and a threshold function with parameter $0.5$. The module $G$ that produces the gating weight will from here on be referred to as the gate. 

\subsection{Training}\label{ssec:training}

We use a pretrained Reinforcement Learning (RL) expert, trained as in \citet{babyguide} and \citet{babyai} by proximal policy optimization \citep{ppo}. After training, the expert is placed once in many missions in order to create training data containing the expert behavior. Using imitation learning as in \citet{babyguide}, we have a cross-entropy loss function $L_{\text{ce}}$ that measures how much the distribution over actions given by the policy of our model deviates from the ``correct'' action of the RL expert.\footnote{For mitigating confusion, we mention explicitly that the RL expert is not the same as the guide: the expert creates the data that is used for backpropagating the model and thus for training it \emph{following} the choice of the action. The guide, however, gives its guidance \emph{prior to} the decision about the chosen action.} Furthermore, we penalize the learner for asking for guidance by adding the gating weight to the loss and balance these incentives by a hyperparameter $\lambda$:
\begin{equation}
L = L_{\text{ce}} + \lambda \cdot g.
\end{equation}
We use $\lambda = 0.3$ in all GoToObj experiments and $0.05$ for PutNextLocal, values that were found by hyperparameter search. The combined model consisting of pre-trained guide, also trained by imitation learning as in \citet{babyguide}, and the newly initialized learner is then trained end-to-end by backpropagating the gradients to all the weights of the combined model. In order to pass the gradients also through the discrete gate $G$, we use the straight-through estimator \citep{straight-through, binarized_straight_through}. In order to allow the learner to learn the usefulness of the guidance at the beginning of training, we initialize $G$ with a positive bias.

\section{Experiments}

\begin{figure*}[ht!]
\centering
\begin{minipage}[t]{0.49\linewidth}
    \centering
    \includegraphics[width=\linewidth]{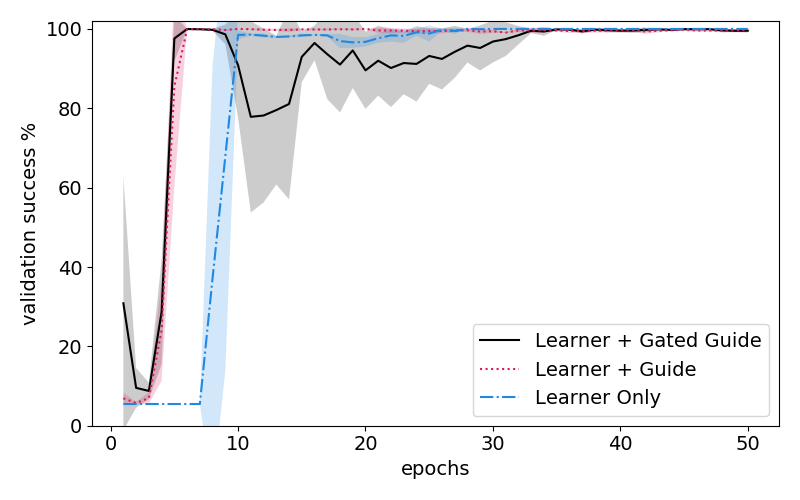}
    \caption*{(a) Baseline Success Rate Comparison}
    \label{fig:base_comp}
\end{minipage}
\begin{minipage}[t]{0.49\linewidth}
     \centering
    \includegraphics[width=\linewidth]{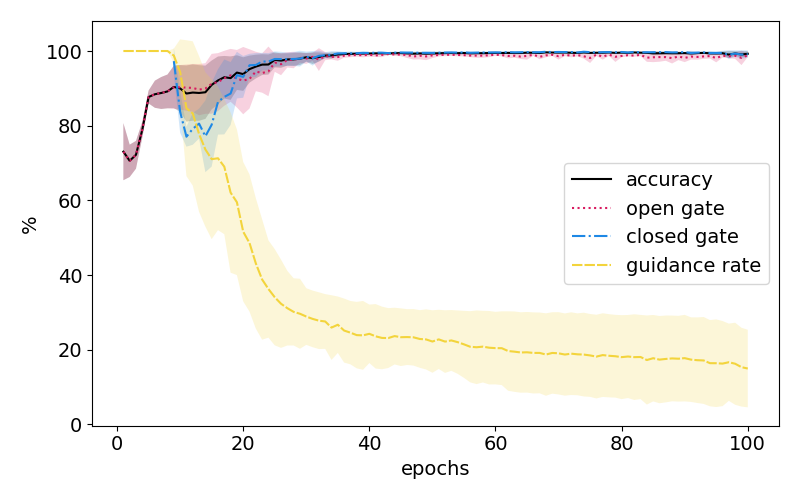}
    \caption*{(b) Validation Accuracy Comparison}
     \label{fig:acc-comp}
\end{minipage}%
 \caption{Results of training our combined model until convergence on GoToObj. Results are averaged over 7 runs. Shaded regions represent standard deviations.}%
 \label{plots}%
\end{figure*}

\begin{figure}[ht!]
\centering
\includegraphics[width=\linewidth]{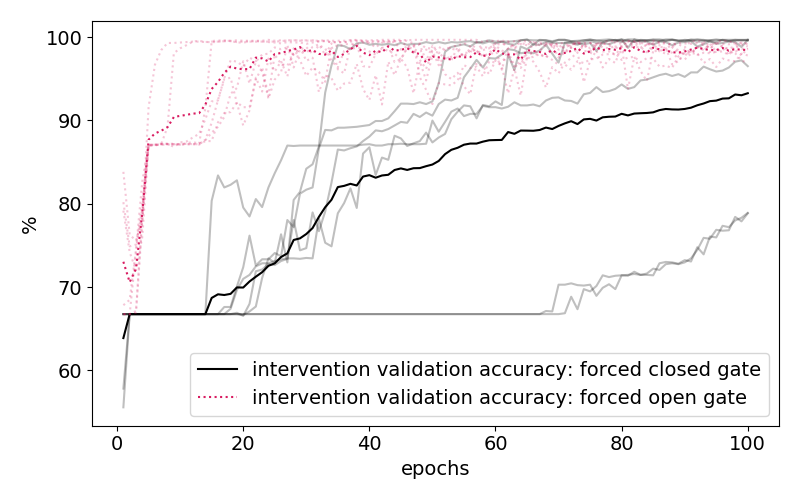}
\caption{We compare the accuracy during validation in cases of forced open and closed gates: irrespective of the gating weight $g_t$ computed from the system, we set $g_t = 1$ (so that the policy bases its decision on the encoded guidance $\text{Enc}(m_t)$) for the red dotted curve and $g_t = 0$ for the black curve.}
\label{fig:guided_unguided}
\end{figure}

In this section, we describe the experiments conducted in order to test the setting of a learner that queries the guide for help. In order to assess this, we train the combined model with $\lambda = 0.3$ for $7$ runs on the simplest level, GoToObj, until convergence. In this level, the learner is instructed to go to a specific object. Results with $\lambda = 0.05$ for $7$ runs on the level PutNextLocal can be found in Appendix \ref{more_results_pnl}.

\textbf{Performance and dynamics.} First of all, we are interested in how our model performs compared to baselines. The first baseline is the learner on its own trained with imitation learning, which was the setting in \citet{babyai}. The second baseline is the learner that receives guidance in each round without the need to query for it, which was studied in \citet{babyguide}. We expect that our model learns faster than the original learner model, due to its access to guidance, but slower than the guided learner without the gate, since it does not always receive guidance. The results can be found in Figure \ref{plots} (a), where we plot the success rate, i.e. portion of successfully completed episodes, during validation.

We monitor the average gating weight over epochs, which we call \textit{guidance rate}, to inspect the usage of the gate over the course of the training. Furthermore, we compare the overall accuracy with the accuracy conditioned on the cases of an open or closed gate to assess the influence on performance. These metrics can be found in Figure \ref{plots} (b)\footnote{The accuracy is the percentage of chosen actions that coincide with the ``correct'' action of the RL expert that is used in the imitation learning process.}.

\begin{figure}
\centering
\includegraphics[width=\linewidth]{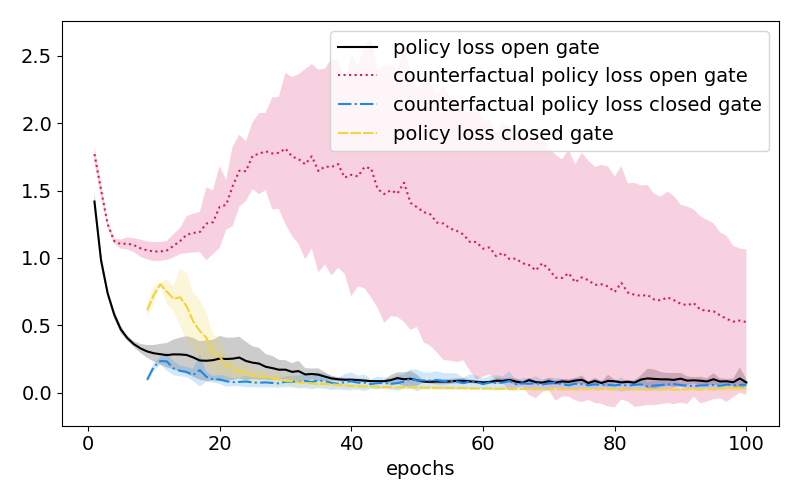}
\caption{Policy Loss Comparison}
\label{fig:pol-loss}
% Behind the vertical line, we see the actual gating decision of the learner. Irrespective of what it wants to do, for each of the plots we intervene in the gating weight $g_t$ and set it to either $0$ or $1$.
\end{figure}

Since the accuracy plots in Figure \ref{plots} (b) are solely correlational, we furthermore plot the validation accuracy for the case where we \emph{intervene} during validation in the gate in order to have it opened or closed to assess the causal influence of the open gate on the accuracy, see Figure \ref{fig:guided_unguided}. 

\textbf{Economic requests.} While the accuracy measures how often the agent was ``right'', the cross-entropy policy loss gives greater insights into the performance with respect to the actual training objective. We would like to assess whether the learner uses the gate economically, since it is penalized. This means to ask for guidance in situations in which it expects the greatest reduction in the policy loss. 
The policy loss for cases of open and closed gate can therefore be found in Figure \ref{fig:pol-loss}. We compare it with the counterfactual policy losses that arise if we force the gate to be opened if the learner wants it to be closed and vice versa. This intervention now allows us to assess the \emph{causal} influence of the gate on the policy loss.

\begin{figure}[ht!]
\centering
\includegraphics[width=1.0\linewidth]{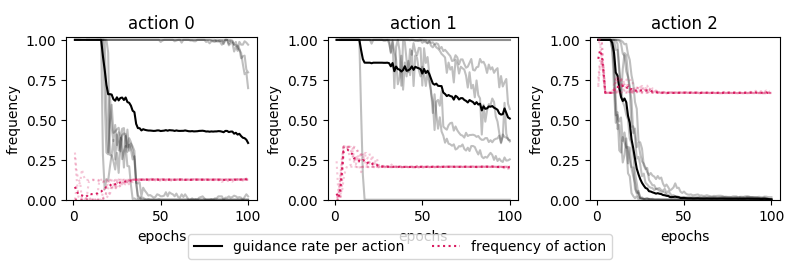}
\caption{Frequency of open gate conditioned on actions and frequencies of actions themselves.}
\label{fig:guidance_per_action}
\end{figure}

\begin{figure}[ht!]
\centering
\includegraphics[width=1\linewidth]{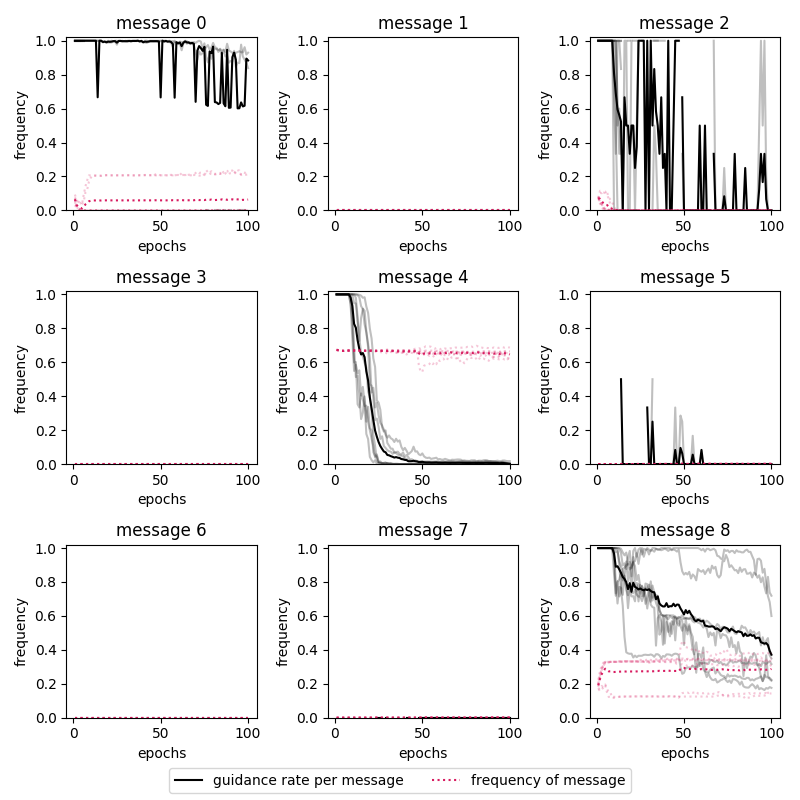}
\caption{Frequency of open gate conditioned on messages and frequencies of messages themselves.}
\label{fig:guidance_per_message}
\end{figure}

\textbf{Guidance semantics.} Finally, we are interested in whether there are meaningful correlations between the frames in which the learner asks for guidance and the actions that the learner takes (Figure \ref{fig:guidance_per_action}) as well as the messages emitted by the guide (Figure \ref{fig:guidance_per_message}). By analyzing this, we can see if the learner masters certain situations that require a certain action or are accompanied by a certain message.

\section{Analysis}\label{sec:Analysis}

\subsection{Results and analysis of experiments}

After outlining the experiments, we now briefly analyze the results. 

\textbf{Performance and dynamics.} First, we notice in Figure \ref{plots} (a) that, while initially our model is indeed faster than the learner baseline and keeps up with the guided baseline, from epoch $10$ to $30$ there is a performance dip not seen in the former models. This is precisely the phase in which the gate successively closes more and more, as can be seen in Figure \ref{plots} (b). Our explanation for this dip is that the policy first needs to adjust to the fact that it does not get the familiar guidance anymore. Eventually, from epoch $30$ onward, our model performs almost perfectly and as well as the baselines.

As mentioned already, the learner becomes indeed more independent over time, roughly choosing its action on its own in $80 \%$ of the cases from epoch $50$ onward. As soon as the guidance rate begins to drop in epoch $10$, we can compute an accuracy conditioned on cases where the gate is closed, as can be seen in the blue line of Figure \ref{plots} (b). This accuracy is lower than the accuracy in case of an open gate (red dotted line) and only fully catches up roughly after $25$ epochs of training.

About the intervention accuracy in Figure \ref{fig:guided_unguided}: we observe that the initial phase with a guidance rate at $1$ sees a steep increase in accuracy with guidance but nearly no change in the accuracy without guidance. We suppose that is the case since in this phase the training happens exclusively with an open gate. Then, the guidance rate drops and training happens increasingly without guidance. Accordingly, the accuracy without guidance starts to increase and eventually catches up. In many individual runs, the performance without guidance is ultimately actually even better. We hypothesize that this is since the gate is mostly closed and so the policy doesn't ``expect'' the gate to be open anymore. Consequently, an open gate and additional encoded message is confusing and leads to misbehavior. Intuitively, this is similar to how humans who are very experienced in their profession may actually just be distracted by someone who occasionally tries to give them advice instead of just letting them do their task on their own.

\textbf{Economic requests.} In order to get a better feeling for how smart the agent is in its guidance requests, we look at Figure \ref{fig:pol-loss}. For similar results about the entropy, see Appendix \ref{more_results}. We see an overall tendency for the policy loss to decrease, as we would expect due to the training. At the same time, the situations where the gate is open are those that are more difficult for the learner (including in the comparison of the counterfactual cases where we change the gate). Additionally, in those situations the reduction in policy loss achieved by asking for guidance is greater -- this can be seen by comparing with the counterfactual situations. We furthermore observe that after the guidance rate starts to drop around epoch $10$, the policy loss in situations of a closed gate rapidly sinks as the learner adapts to those novel situations. In the meantime, the open gate policy loss stabilizes until around epoch $20$, while the counterfactual policy loss in these situations strongly increases. This indicates that the learner learns to selectively open the gate in situations that are more difficult and especially so without guidance. 

\textbf{Guidance semantics.} To gain insights about dependencies between specific actions and the guidance rate, we now look at Figure \ref{fig:guidance_per_action}. We see that in situations where the learner takes action $2$, which corresponds to ``move forward'', the guidance rate drops relatively early to $0$. This may be the case since this is the most common and supposedly most easily identifiable action.
For action $0$ that corresponds to ``turn right'' and action $1$ that corresponds to ``turn left'', we see that the guidance rate also decreases, albeit slower and asymmetrically. This may be due to a higher difficulty of distinguishing those actions from each other. Intuitively they are symmetric and it may often be unclear what to do if ``move forward'' is not a promising action. In some runs, the guidance rate drops more for action $0$ and in some more for action $1$. We may attempt to explain this by the learner either learning to request help in situations where action $0$ is one of the promising options (potentially the only one) or learning the same with action $1$. In both cases, it is ensured that situations with confusion between those two actions are encompassed.

On the guidance rate conditioned on messages: as we can see in Figure \ref{fig:guidance_per_message}, mainly three messages are used to convey guidance and for all of them the guidance rate decreases over time. We suppose that the overall trends happen due to the close correspondence between messages and actions that was already observed by \citet{babyguide}.

\subsection{Guidance in space and time} \label{space_and_time}

So far, we mainly discussed ``global'' metrics, in the sense that we aggregate information over complete epochs. This still leaves open the question how guidance requests evolve with respect to the position of the agent and temporally during an episode.

For the first aim, we create heatmaps as in Figure \ref{heatmap}. For more maps, see Figure \ref{all_heatmaps_gotoobject} in Appendix \ref{more_results}.

\begin{figure}[ht!]
\centering
\begin{minipage}[b]{0.49\linewidth}
    \centering
    \includegraphics[width=1.0\linewidth]{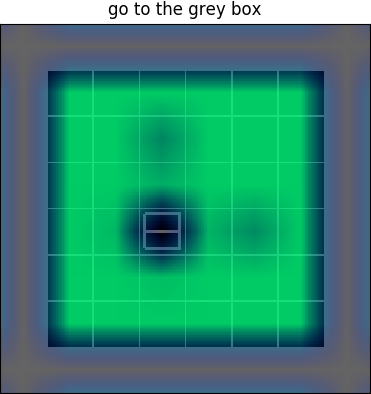}
\end{minipage}
\begin{minipage}[b]{0.49\linewidth}
    \centering
    \includegraphics[width=1.0\linewidth]{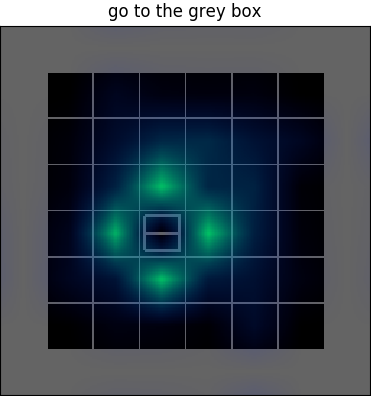}
\end{minipage}
\caption{Two example heatmaps from training in level GoToObj, one in the beginning of training and one in an advanced stage. These heatmaps are created as follows: after an epoch is finished, the agent is placed in a specific mission. Then, we let it follow its path until the episode is over. For each point in the agent trajectory we record whether it asks for guidance. Multiple trajectories are sampled by randomly placing the agent in a new position. The brightness of the color in the figure depicts the average guidance rate within that position.}
\label{heatmap}
\end{figure}

As we can see, the trained agent asks for guidance often specifically if it is near the goal object in order to find out if it should ``turn towards it'', which would cause the goal to be reached. It is important for the agent to be reasonably sure about the goal being reached beforehand, since otherwise turning to the object will result in two lost moves.

%We also see that the agent asks for guidance in this specific example when it is enclosed by many objects. One explanation is that the agent has few actions that walk it out of this situation and therefore needs help. Another possibility is that it asks for help since it is near a grey key, which shares features with the goal and may therefore be ``distracting''.

\begin{figure}[ht!]
\centering
\begin{minipage}[b]{0.47\linewidth}
    \centering
    \includegraphics[width=1.0\linewidth]{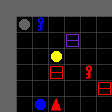}
\end{minipage}
\begin{minipage}[b]{0.46\linewidth}
    \centering
    \includegraphics[width=1.0\linewidth]{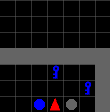}
\end{minipage}
\caption{Observation types in GoToLocal. GoToLocal is a level different from GoToObj or PutNextLocal and is used to illustrate some of the extra possibilities ((1,$x$) and ($x$,1)) on further levels. In the left mission, the agent is tasked to go to the blue ball. This is directly left from the agent, whereas to the right, there is no feature in common with the goal. Therefore, the observation type is $(2,0)$. In the right mission, the goal is to go to a grey ball. Since the blue ball shares one feature with the grey one (namely being a ball), the corresponding observation type is $(1,2)$.}
\label{distractingness}
\end{figure}

\begin{figure}[ht!]
\centering
\includegraphics[width=1.0\linewidth]{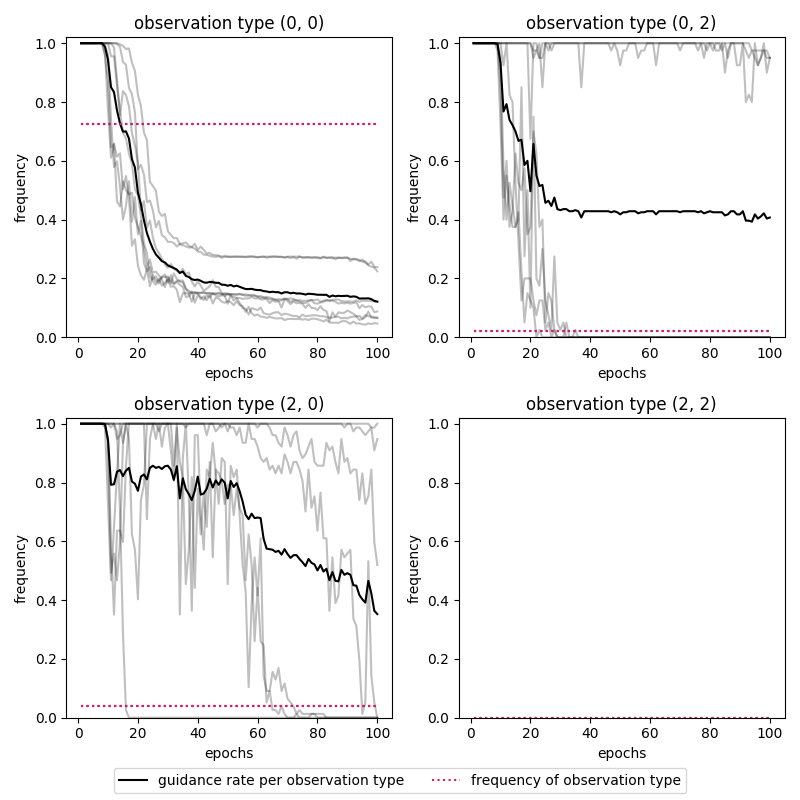}
\caption{Frequency of open gate conditioned on observation types and frequencies of observation types themselves. For example, type $(2,0)$ is a situation where directly left of the agent there is the goal and right of it there is no object sharing any feature with the goal-object. In GoToObj there is only one object in the level so situations like $(1,x)$ or $(x,1)$ do not occur and are left out in this plot.}
\label{guidance_per_observation_type}
\end{figure}

In order to test the influence of objects on the agent more quantitatively, we create another metric that conditions the guidance rate on how ``goal-like'' the observation is that the agent faces. For this matter, we assign tuples $(d_1, d_2)$ to each observation, where $d_1$ and $d_2$ signify how goal-like the object left and right from the agent is. We measure this by the number of features it has in common with the goal object, where features are both color and object-type. $d_i = k$ means that the object shares $k \in \{0, 1, 2\}$ features with the goal object. This creates $9$ observation types\footnote{Note that even the combination $(2, 2)$ is in principle possible in higher levels: There are tasks such as ``Go to a red ball'' where several red ball can be in the mission.  However, this is unlikely and the expert never places itself between two goal objects. Furthermore, in GoToObj there is simply just one object in the mission. Therefore, the graph in Figure \ref{guidance_per_observation_type} corresponding to $(2, 2)$ is empty.}. See Figure \ref{distractingness} for examples. The results can be found in Figure \ref{guidance_per_observation_type}.
% and show the following: when there are distracting objects to the left and right of the agent which are \emph{not} the goal object, then the guidance rate follows precisely the normal trend we already saw in Figure \ref{plots} (b). 
 We can observe strong changes in the guidance rate if the goal object is to the left or right: if it is on the left, the guidance rate is significantly greater and if it is to the right, the guidance rate is significantly smaller than usually. This is in line with the plots of the guidance rate conditioned on actions, Figure \ref{fig:guidance_per_action}, which already showed that turning to the left requires considerably more guidance than turning to the right. This indicates that the high guidance rate at goal objects may to a large extend be caused by the high correlation between turning actions and guidance rates and be mostly independent of the fact that there is a goal around.

Now we turn to the question about the guidance rate in time: within one episode, are there usually phases where more or less guidance is needed? The heatmaps suggest that the agent mostly asks for guidance in the end of the episode.

\begin{figure}[ht!]
\centering
\includegraphics[width=1.0\linewidth]{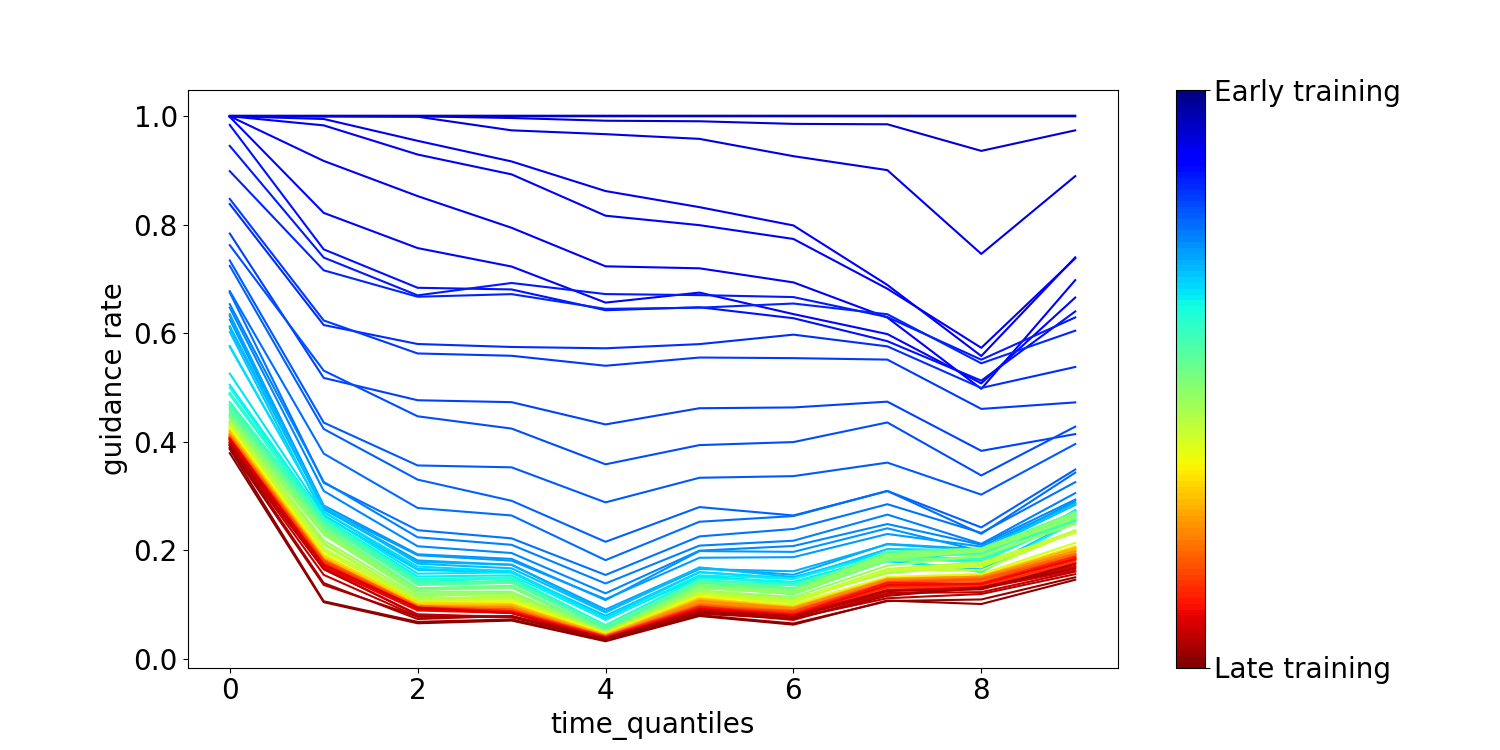}
\caption{Guidance per time quantile: roughly speaking, a timepoint $t$ is in quantile $k$ of $10$ if $t / l \approx k / 10$, where $l$ is the length of the corresponding episode. The plots show the guidance rate corresponding to the different quantiles. Dark blue curves belong to earlier epochs whereas red curves belong to later epochs.}
\label{guidance_per_bucket}
\end{figure}

In order to answer this question, we create the ``guidance per time quantile'' plot,  Figure \ref{guidance_per_bucket}. As we can see, the guidance rate is in general high in the beginning of episodes and drops once more knowledge about the environment is acquired. However, in the end of the episode, the guidance rate grows again and is greatest in the very end, which is in line with the qualitative assessment from the heatmaps.

One interpretation for this is the following: in the beginning, the agent needs to roughly figure out ``in which direction to head''. Once this is clear, it can walk there without further guidance. But in the very end, it needs more precision and asks for guidance again in order to finally find its goal. This is similar to how humans often look at a map in the beginning of a hike in order to figure out the direction, and then in the end again in order to reassess how their new position now relates to the goal.

\section{Conclusion and future research}
In this paper, we extended a recently proposed method to guide a learner in a gridworld environment by letting the learner explicitly ask for help. To accomplish this, we defined a binary gate in the learner's model. We brought the original approach closer to the real world by (i) enabling bi-directional communication and (ii) attaching a cost to it. 

We showed that the learner successfully learns to utilize the guidance gate to achieve a favorable trade-off between learning speed and amount of guidance requested. Initially using the full guidance to learn faster than a learner without guidance, it eventually learns to request guidance only where it is especially helpful, acting increasingly independent.

Future research could consist of retraining the guide to see if it learns to send more abstract messages that provide guidance over multiple time steps. This may be fruitfully combined with giving the guide an additional information advantage like a bird's eye's view so that the guide has more foresight than the learner. It would require a way for the learner to memorize past messages. 

Another direction is to replace the gate by an emergent communication channel from the learner to the guide, so that the learner can send its guidance requests in more nuanced ways. Furthermore, we saw that the policy may have problems dealing with the additional guidance it receives unexpectedly in late training. It may be worthwhile to experiment with policy architectures that can deal better with spontaneous changes in its input.

%Finally, research might as well aim at finding settings where the guide can be meaningfully replaced by a human to allow, in the best case, for learning in terrains that are untraversed by unguided AI systems.

Finally, research might as well aim at finding ways to meaningfully replace the guide agent by a human. This might allow for better learning in tasks that autonomous agents struggle to learn by themselves.

\bibliography{library}
\bibliographystyle{acl_natbib}

\clearpage

\appendix

\section{More results for Gotoobj}\label{more_results}

We use this part of the appendix to give additional figures for the Gotoobj level.

\begin{figure}[ht!]
\centering
\includegraphics[width=\linewidth]{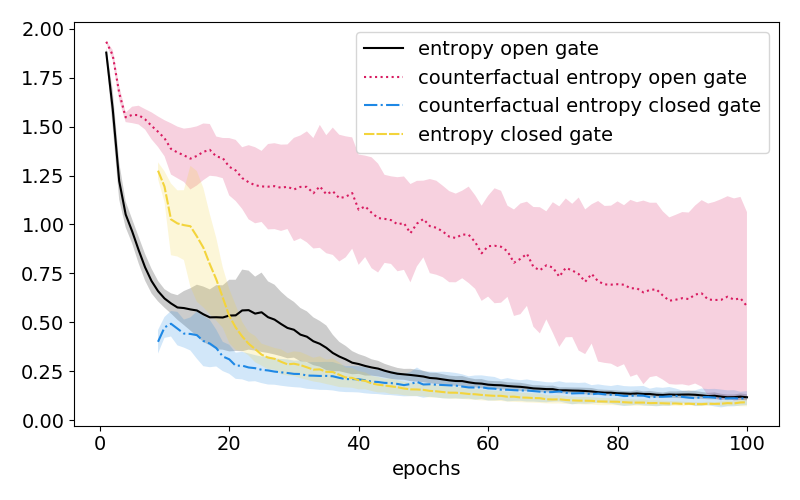}
\caption{Entropy Comparison for GoToObj}
\label{fig:d-entr}
\end{figure}

First, similarly to the reduction in policy loss as investigated in section \ref{sec:Analysis}, we are also interested in the reduction of overall uncertainty (Shannon entropy) over actions that the learner achieves by asking for guidance and compare this again for the cases where the learner actually wants to open or close the gate. This can be found in Figure \ref{fig:d-entr}. Note however the subtlety that the learner may become more certain about which actions to choose, but focus on the wrong action. By spotting differences between the two measures we may identify situations in which the guidance misleads the learner or, vice versa, in which a very certain learner that aims for the wrong action actually gets less certain by getting the correct guidance, leading to a reduction in policy loss.

As we see, the shape of the resulting plots are relatively similar to those of the policy loss. As one notable difference, the counterfactual entropy does not temporarily increase in the same way as the policy loss for open gate situations. This indicates that the learner generally gets continuously more certain, albeit not necessarily about the right actions.
% (This was the end of the analysis of entropy and loss, maybe we can reformulate it so that it fits here!)

\begin{figure*}[ht!]
\begin{subfigure}{0.32\linewidth}
\includegraphics[width=\linewidth]{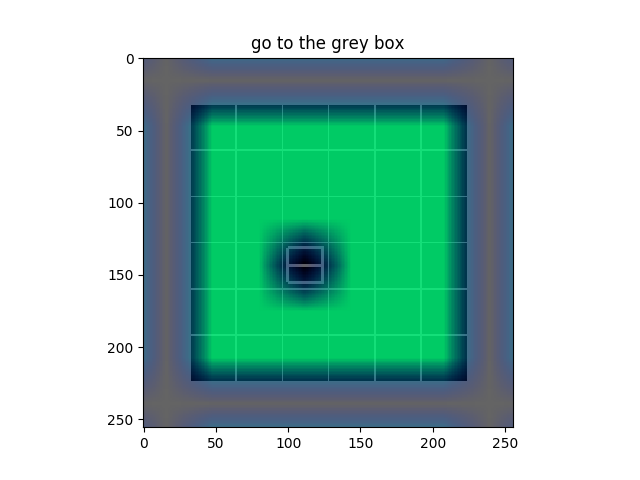}
\end{subfigure}
\begin{subfigure}{0.32\linewidth}
\includegraphics[width=\linewidth]{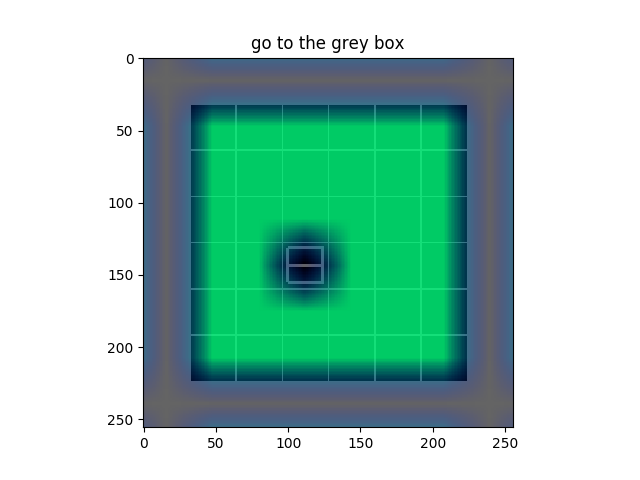}
\end{subfigure}
\begin{subfigure}{0.32\linewidth}
\includegraphics[width=\linewidth]{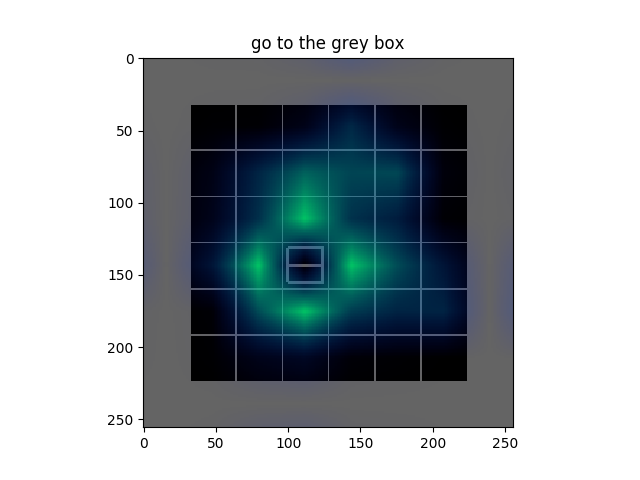}
\end{subfigure}
\begin{subfigure}{0.32\linewidth}
\includegraphics[width=\linewidth]{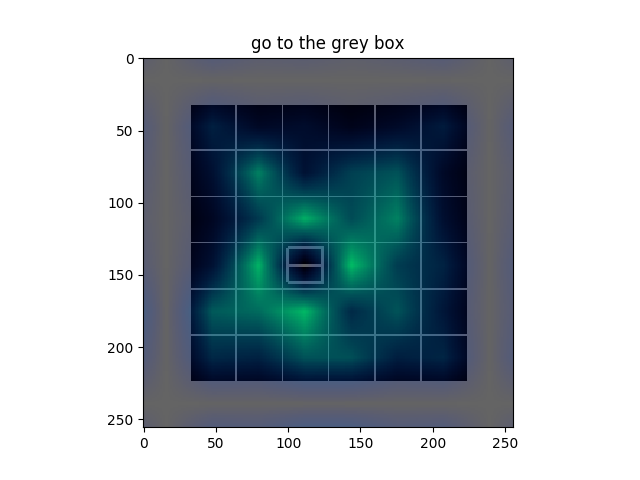}
\end{subfigure}
\begin{subfigure}{0.32\linewidth}
\includegraphics[width=\linewidth]{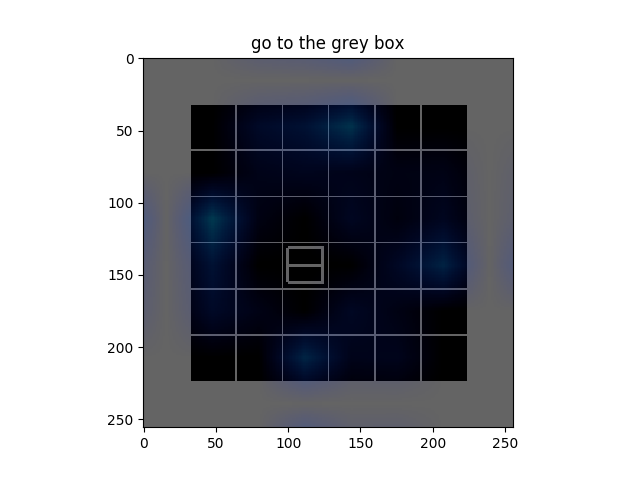}
\end{subfigure}
\begin{subfigure}{0.32\linewidth}
\includegraphics[width=\linewidth]{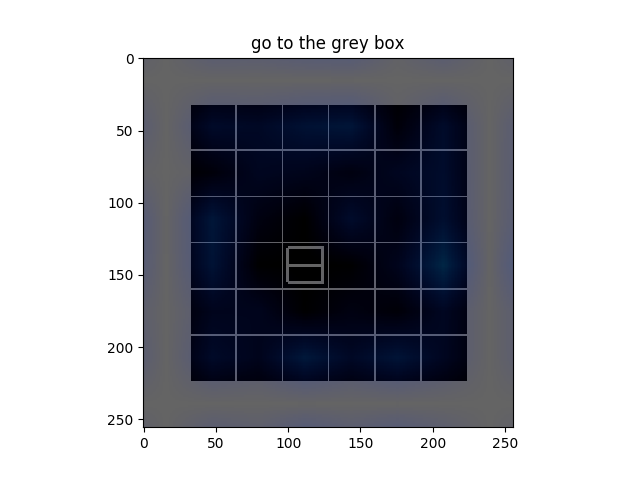}
\end{subfigure}
\begin{subfigure}{0.32\linewidth}
\includegraphics[width=\linewidth]{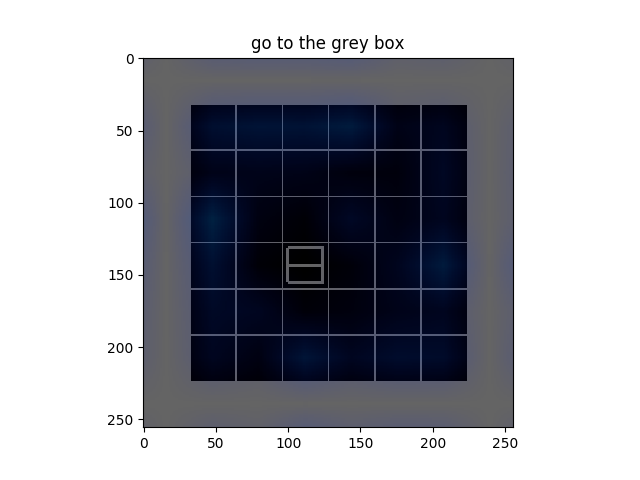}
\end{subfigure}
\begin{subfigure}{0.32\linewidth}
\includegraphics[width=\linewidth]{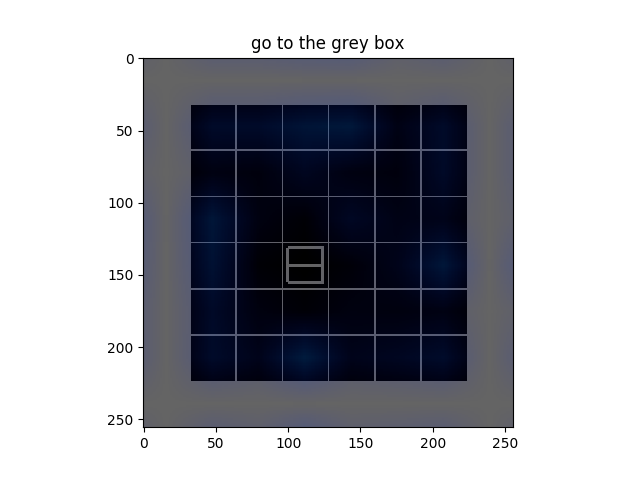}
\end{subfigure}
\begin{subfigure}{0.32\linewidth}
\includegraphics[width=\linewidth]{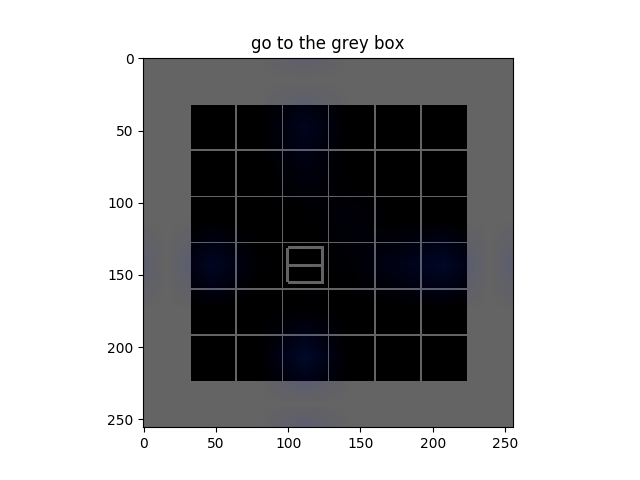}
\end{subfigure}
\begin{subfigure}{0.32\linewidth}
\includegraphics[width=\linewidth]{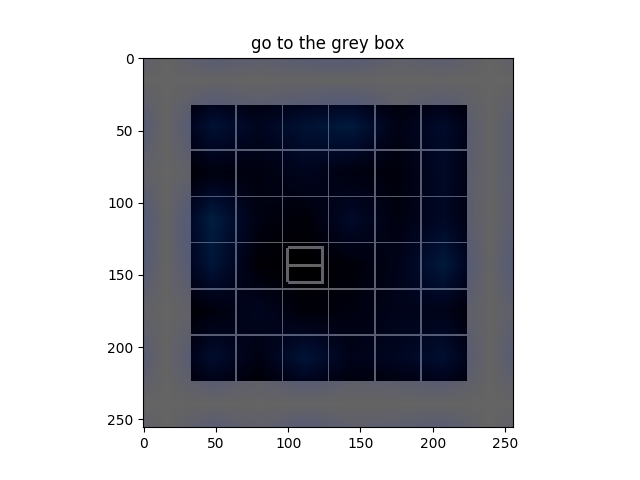}
\end{subfigure}
\begin{subfigure}{0.32\linewidth}
\includegraphics[width=\linewidth]{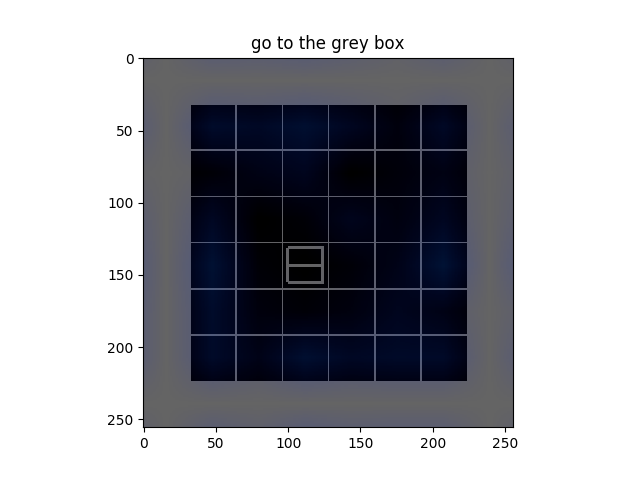}
\end{subfigure}
\begin{subfigure}{0.32\linewidth}
\includegraphics[width=\linewidth]{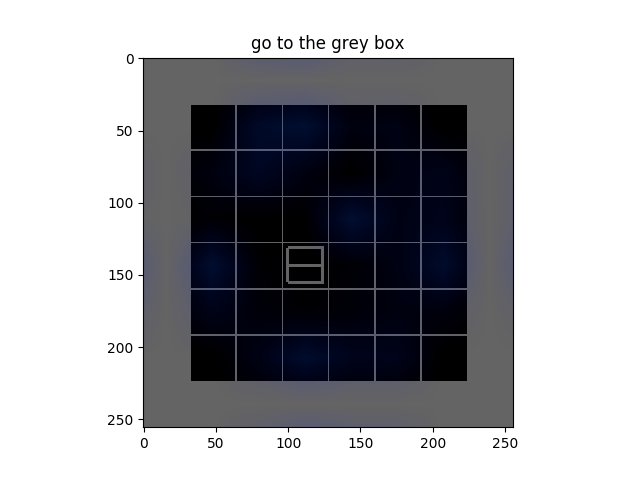}
\end{subfigure}
\begin{subfigure}{0.32\linewidth}
\includegraphics[width=\linewidth]{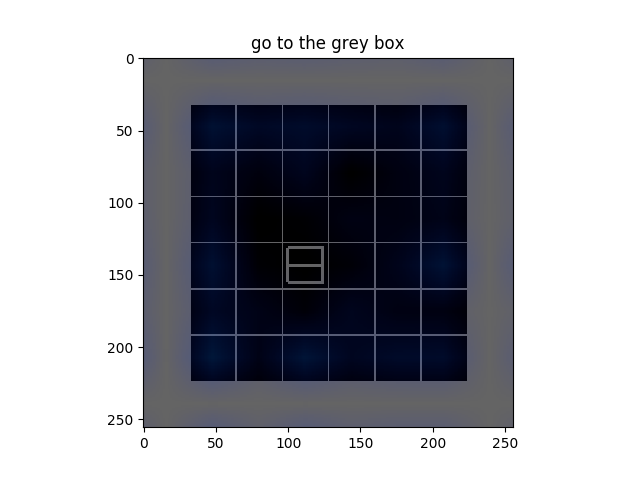}
\end{subfigure}
\begin{subfigure}{0.32\linewidth}
\includegraphics[width=\linewidth]{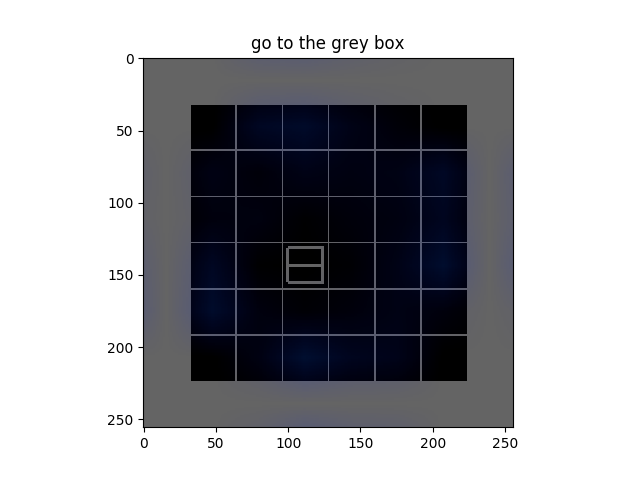}
\end{subfigure}
\begin{subfigure}{0.32\linewidth}
\includegraphics[width=\linewidth]{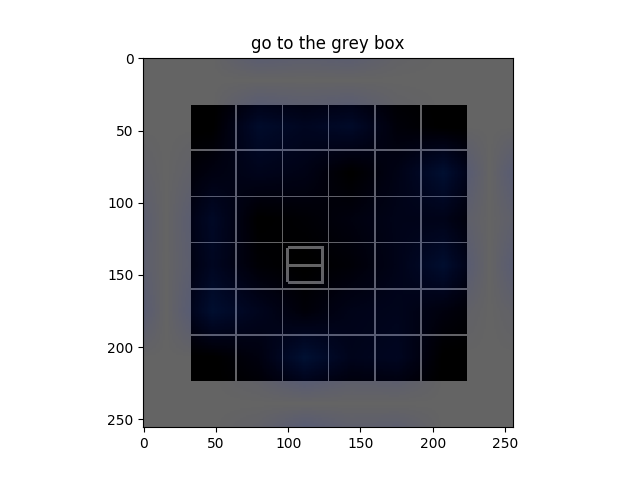}
\end{subfigure}
\begin{subfigure}{0.33\linewidth}
\includegraphics[width=\linewidth]{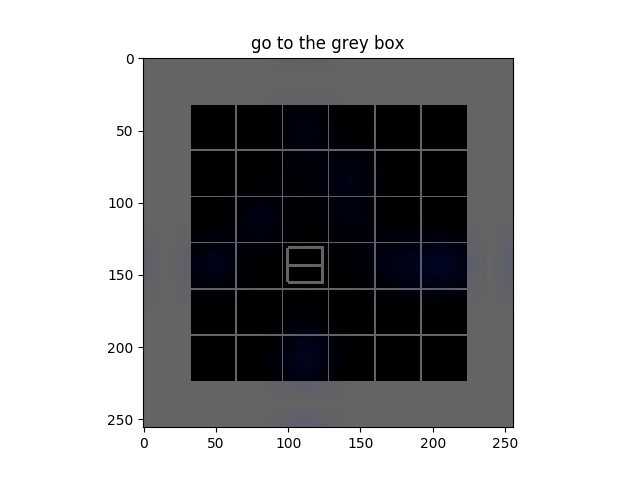}
\end{subfigure}
\begin{subfigure}{0.33\linewidth}
\includegraphics[width=\linewidth]{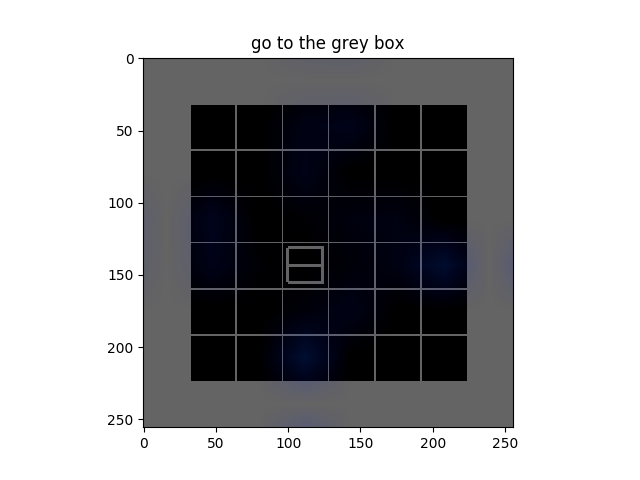}
\end{subfigure}
\begin{subfigure}{0.33\linewidth}
\includegraphics[width=\linewidth]{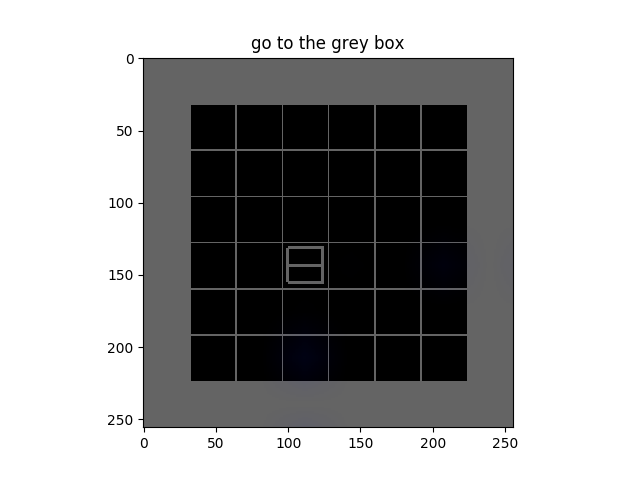}
\end{subfigure}
\caption{Heatmaps for GoToObj. They are ordered from left to right and then top to bottom. This shows how the guidance requests evolve over the course of the whole training in one specific example mission. Over time, not much guidance remains.}
\label{all_heatmaps_gotoobject}
\end{figure*}

Second, in figure \ref{all_heatmaps_gotoobject} we give a more detailed view of the development of the spatial frequency of guidance requests as discussed in section \ref{space_and_time}.

\clearpage

\clearpage

\section{Results for Putnextlocal} \label{more_results_pnl}
%%%%%%%%%%%%%%%%%%%%%%%%%%%%%%%%%% PutNextLocal %%%%%%%%%%%%%%%%%%
%%%%%%%%%%%%%%%%%%%%%%%%%%%%%%%%%
We provide results for PutNextLocal corresponding to the same results for GoToObj which were in the body of the paper. As is visible, many of our findings remain valid in this higher level. Note however that over the course of training, we observe significant overfitting. We nevertheless showed the whole development of the learning process in order to show the full development of the guidance rate.

\begin{figure*}
\centering
\begin{minipage}[t]{0.49\linewidth}
    \centering
    \includegraphics[width=\linewidth]{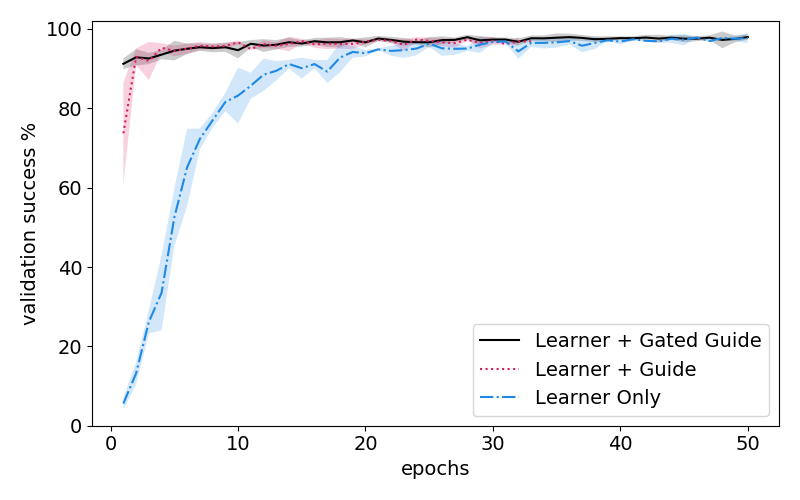}
    {(a) Baseline comparison} 
\end{minipage}
\begin{minipage}[t]{0.49\linewidth}
     \centering
    \includegraphics[width=\linewidth]{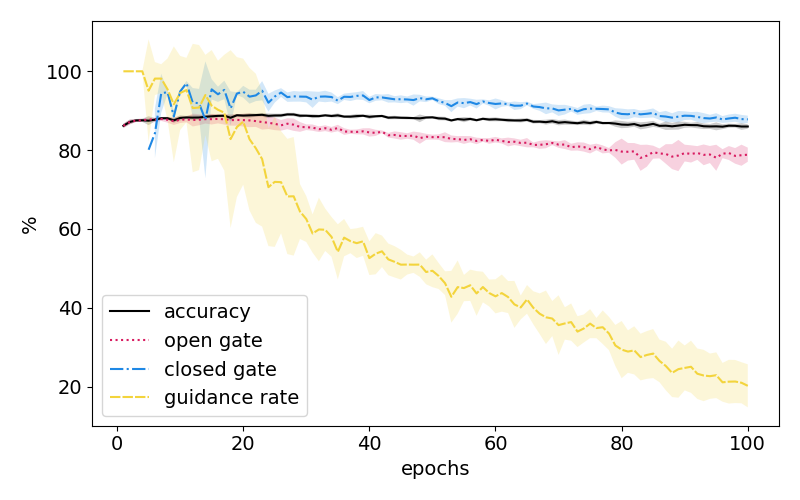}
    {(b) Validation Accuracy Comparison} 
\end{minipage}%

\begin{minipage}[t]{0.49\linewidth}
    \centering
    \includegraphics[width=\linewidth]{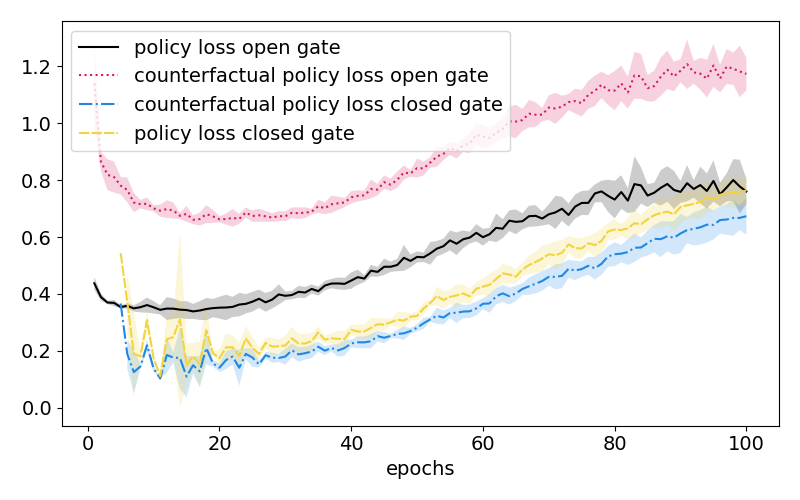}
    {(c) Policy Loss comparison} 
    % Behind the vertical line, we see the actual gating decision of the learner. Irrespective of what it wants to do, for each of the plots we intervene in the gating weight $g_t$ and set it to either $0$ or $1$.
\end{minipage}
\begin{minipage}[t]{0.49\linewidth}
    \centering
    \includegraphics[width=\linewidth]{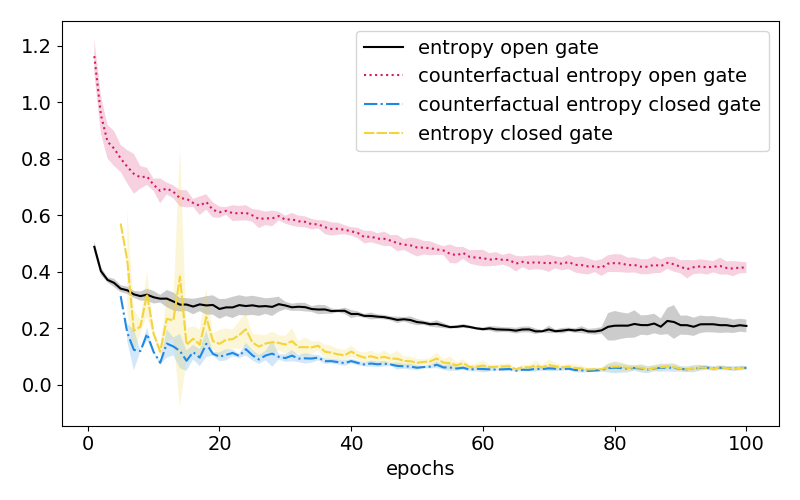}
    {(d) Entropy comparison}
\end{minipage}%
\caption{PutNextLocal}
\end{figure*}

\begin{figure}
\centering
\includegraphics[width=1\linewidth]{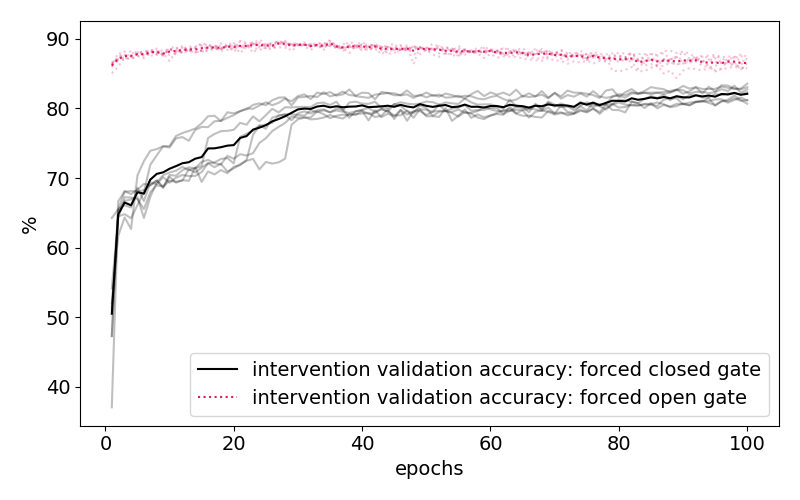}
\caption{PutNextLocal: We compare the accuracy during validation in cases of forced open and closed gates: irrespective of the gating weight $g_t$ computed from the system, we set $g_t = 1$ (so that the policy bases its decision on the encoded guidance $\text{Enc}(m_t)$) for the red dotted curve and $g_t = 0$ for the black curve.}
\end{figure}

\begin{figure}
\centering
\includegraphics[width=1.0\linewidth]{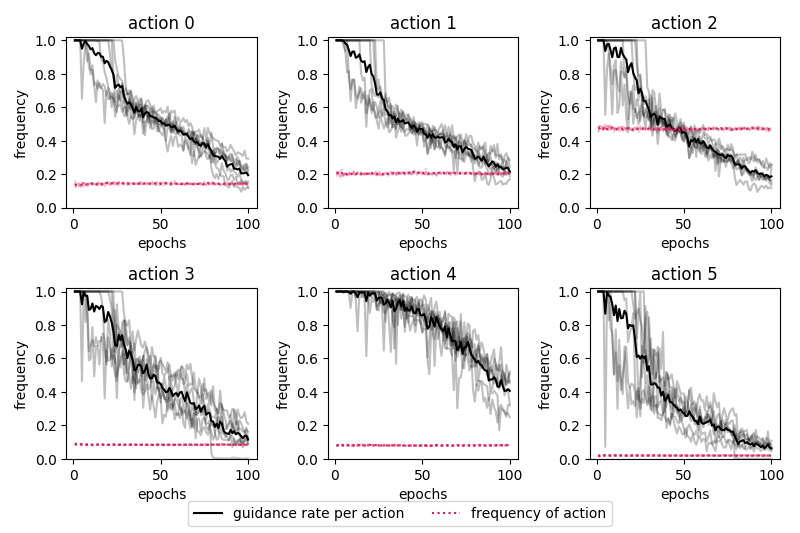}
\caption{PutNextLocal: Frequency of open gate conditioned on actions and frequencies of actions themselves.}
\end{figure}

\begin{figure}
\centering
\includegraphics[width=1\linewidth]{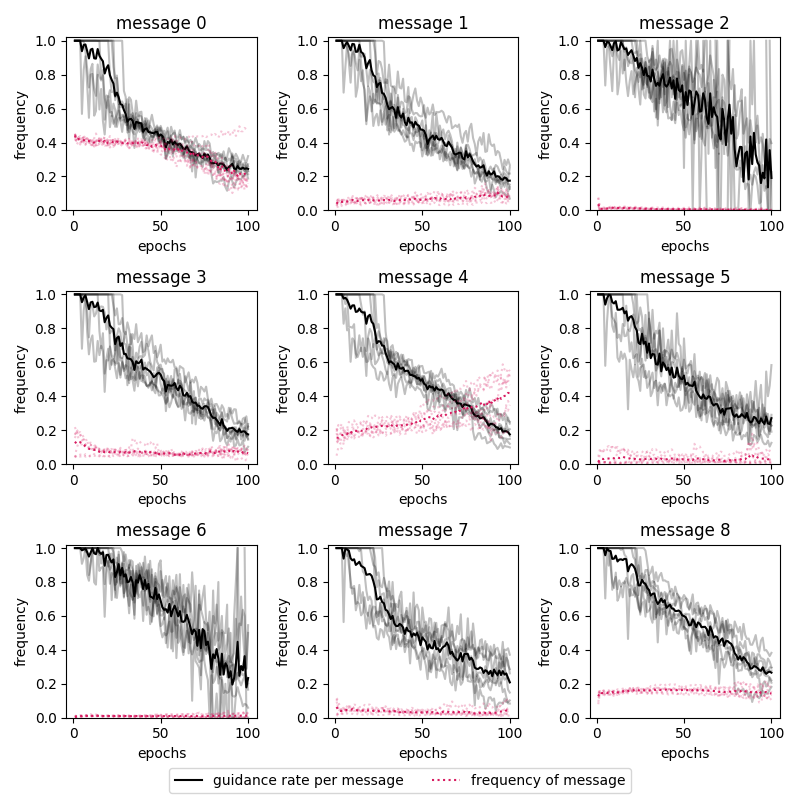}
\caption{PutNextLocal: Frequency of open gate conditioned on messages and frequencies of messages themselves.}
\end{figure}

\begin{figure}
\centering
\includegraphics[width=1.0\linewidth]{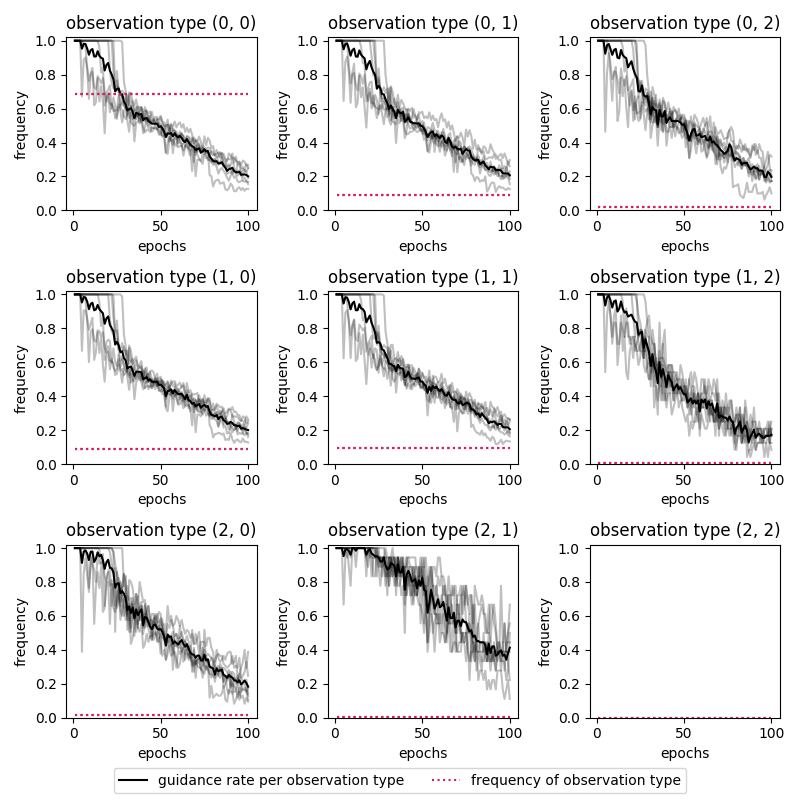}
\caption{PutNextLocal: Frequency of open gate conditioned on observation types and frequencies of observation types themselves. For example, type $(2,1)$ is a situation where directly left of the agent there is the goal and right of it there is an object sharing one feature with the goal-object.}
\end{figure}

\begin{figure}[t!]
\centering
\includegraphics[width=1.0\linewidth]{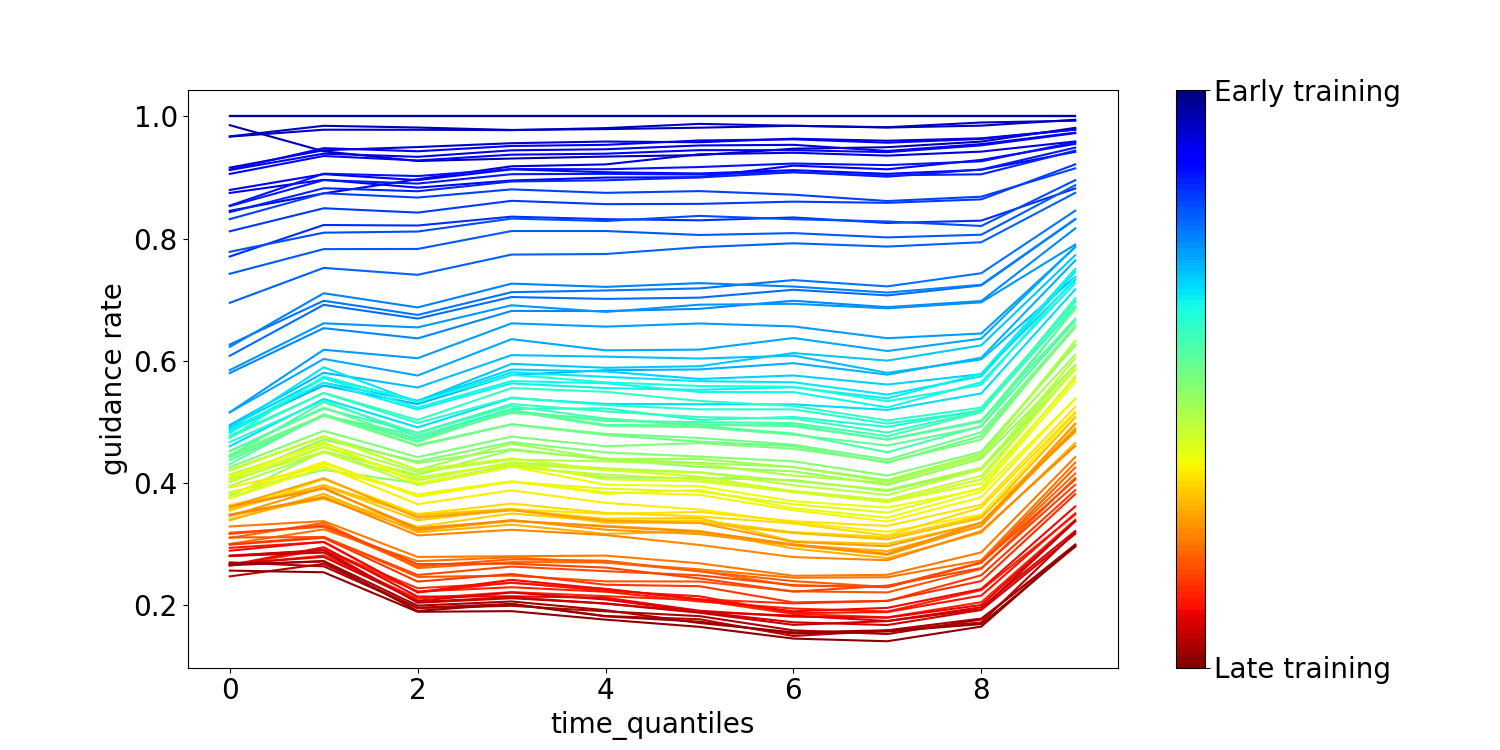}
\caption{PutNextLocal: Guidance per time quantile: roughly speaking, a timepoint $t$ is in quantile $k$ of $10$ if $t / l \approx k / 10$, where $l$ is the length of the corresponding episode. The plots show the guidance rate corresponding to the different quantiles. Dark blue curves belong to earlier epochs whereas red curves belong to later epochs.}
\end{figure}

\clearpage

\clearpage
\begin{figure*}[ht!]
\begin{subfigure}{0.32\linewidth}
\includegraphics[width=\linewidth]{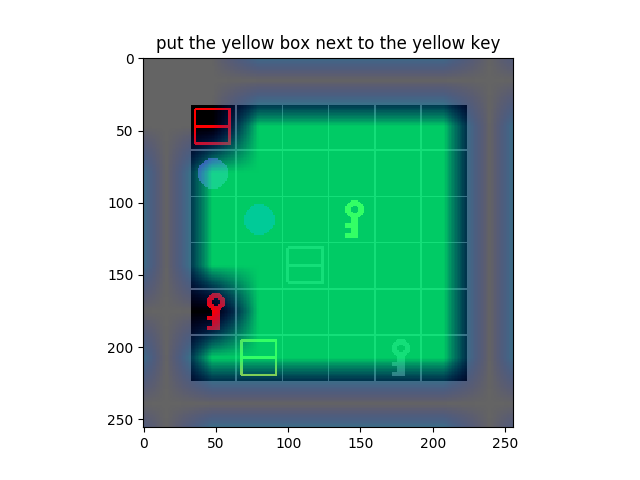}
\end{subfigure}
\begin{subfigure}{0.32\linewidth}
\includegraphics[width=\linewidth]{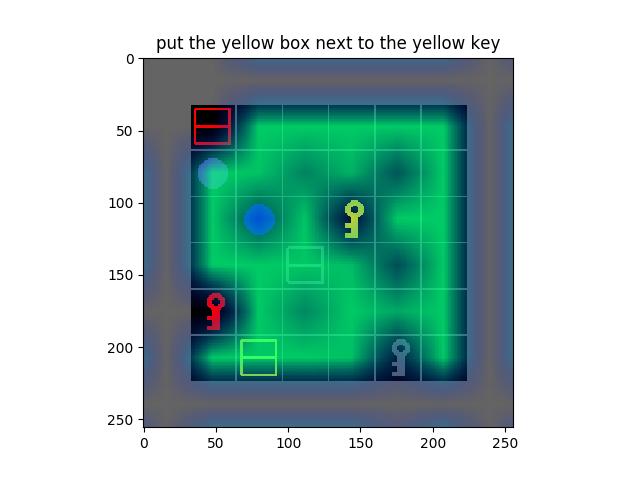}
\end{subfigure}
\begin{subfigure}{0.32\linewidth}
\includegraphics[width=\linewidth]{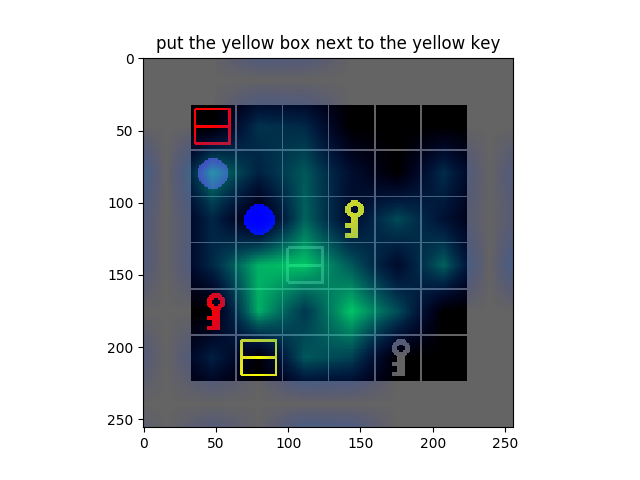}
\end{subfigure}
\begin{subfigure}{0.32\linewidth}
\includegraphics[width=\linewidth]{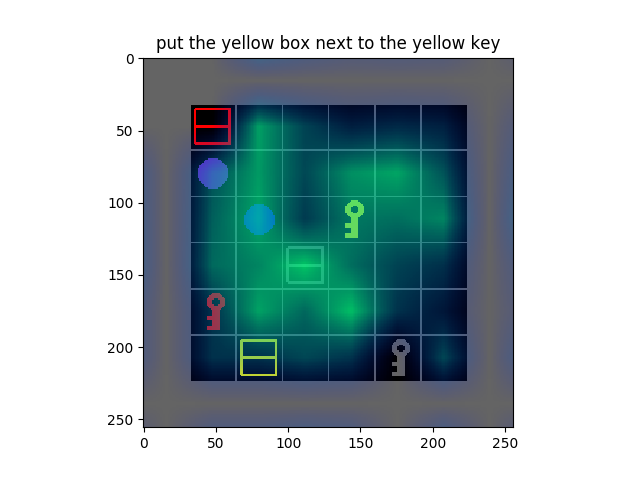}
\end{subfigure}
\begin{subfigure}{0.32\linewidth}
\includegraphics[width=\linewidth]{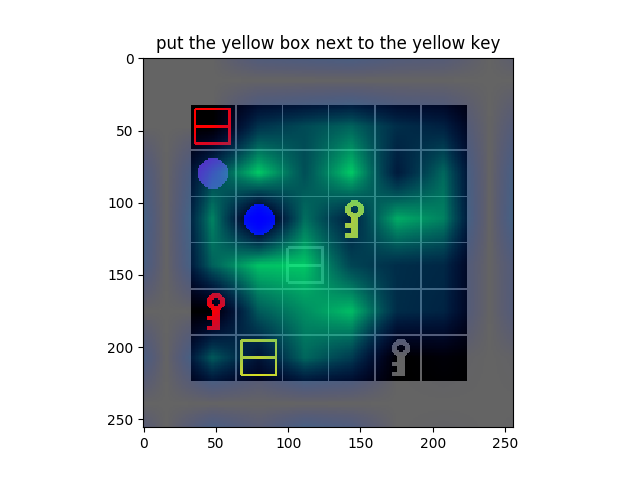}
\end{subfigure}
\begin{subfigure}{0.32\linewidth}
\includegraphics[width=\linewidth]{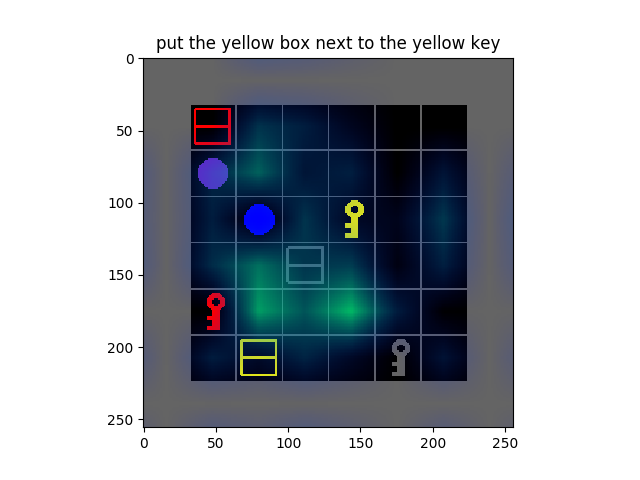}
\end{subfigure}
\begin{subfigure}{0.32\linewidth}
\includegraphics[width=\linewidth]{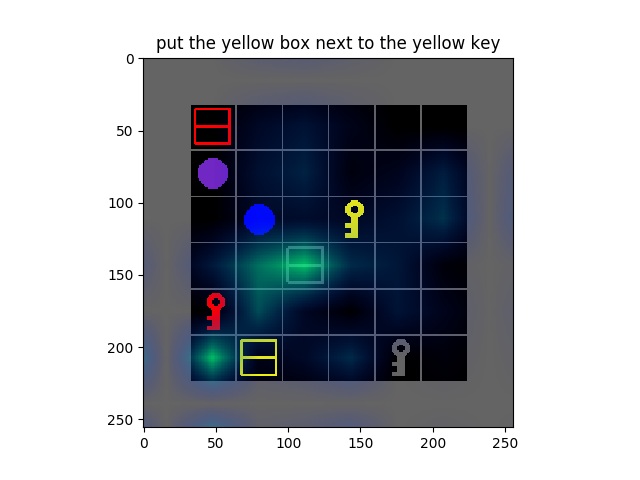}
\end{subfigure}
\begin{subfigure}{0.32\linewidth}
\includegraphics[width=\linewidth]{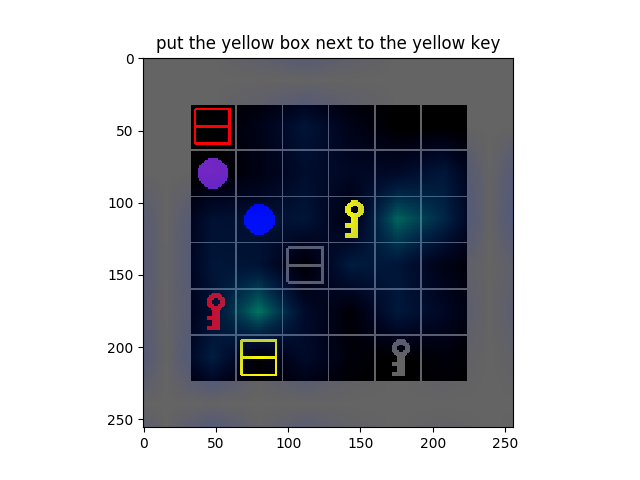}
\end{subfigure}
\begin{subfigure}{0.32\linewidth}
\includegraphics[width=\linewidth]{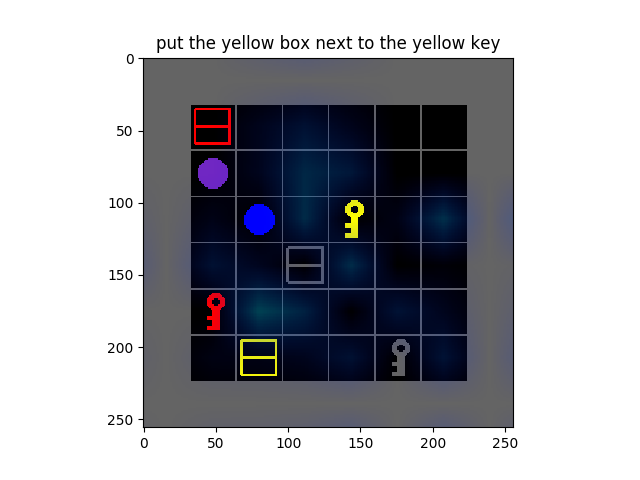}
\end{subfigure}
\begin{subfigure}{0.32\linewidth}
\includegraphics[width=\linewidth]{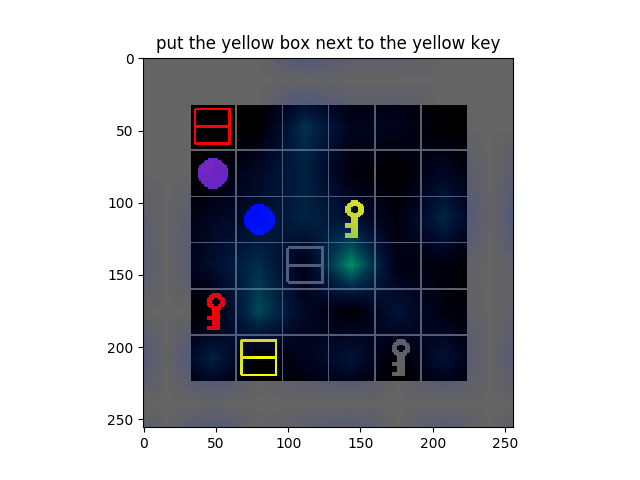}
\end{subfigure}
\begin{subfigure}{0.32\linewidth}
\includegraphics[width=\linewidth]{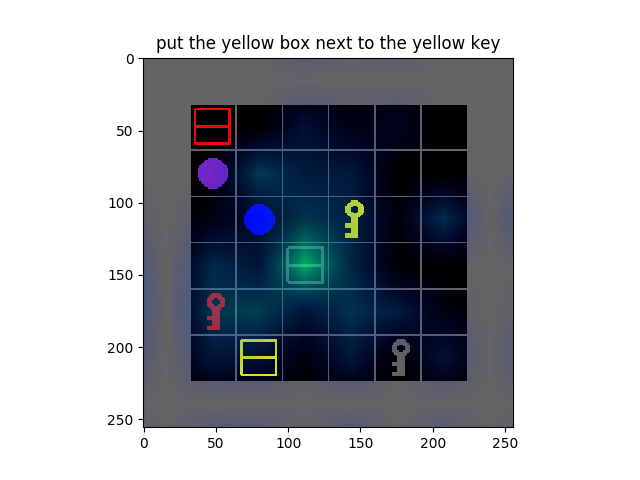}
\end{subfigure}
\begin{subfigure}{0.32\linewidth}
\includegraphics[width=\linewidth]{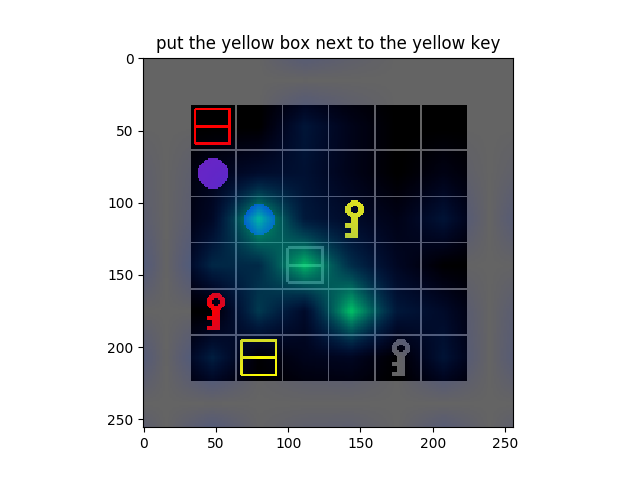}
\end{subfigure}
\begin{subfigure}{0.32\linewidth}
\includegraphics[width=\linewidth]{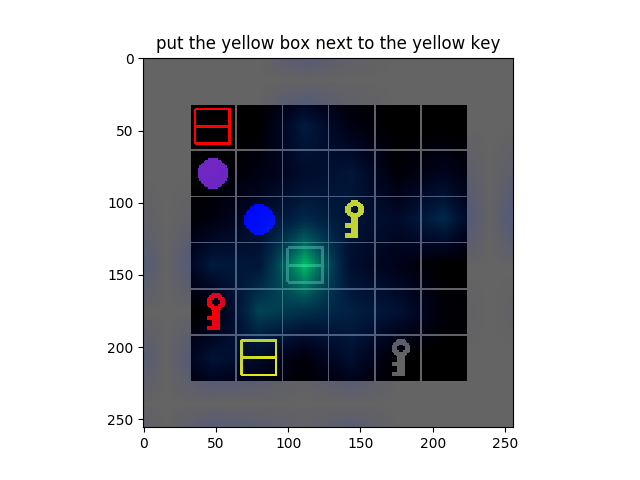}
\end{subfigure}
\begin{subfigure}{0.32\linewidth}
\includegraphics[width=\linewidth]{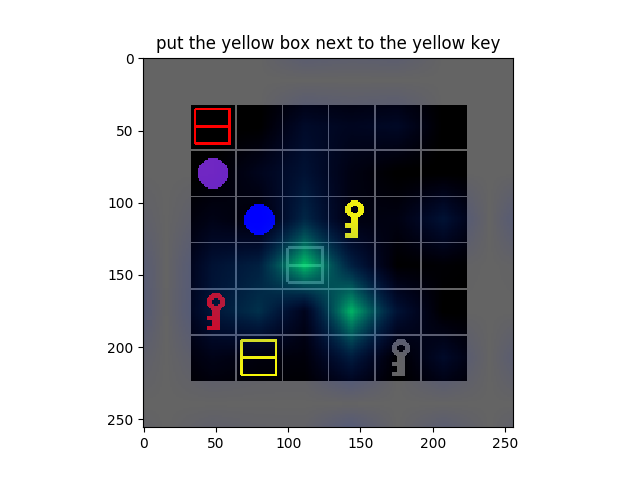}
\end{subfigure}
\begin{subfigure}{0.32\linewidth}
\includegraphics[width=\linewidth]{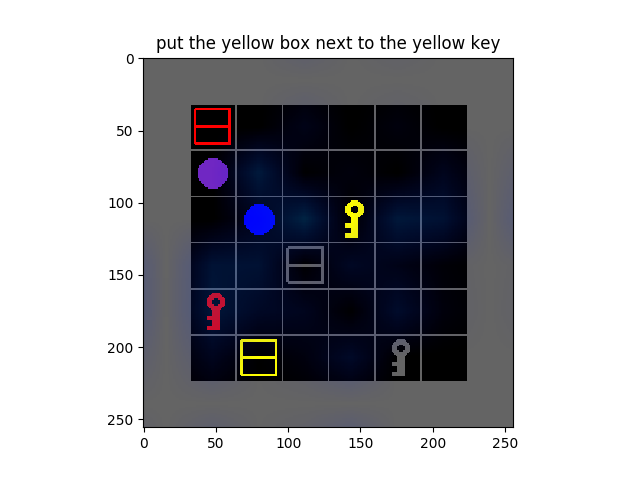}
\end{subfigure}
\begin{subfigure}{0.33\linewidth}
\includegraphics[width=\linewidth]{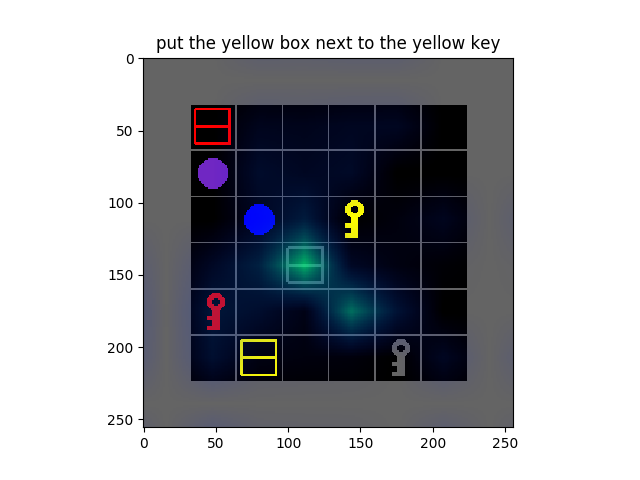}
\end{subfigure}
\begin{subfigure}{0.33\linewidth}
\includegraphics[width=\linewidth]{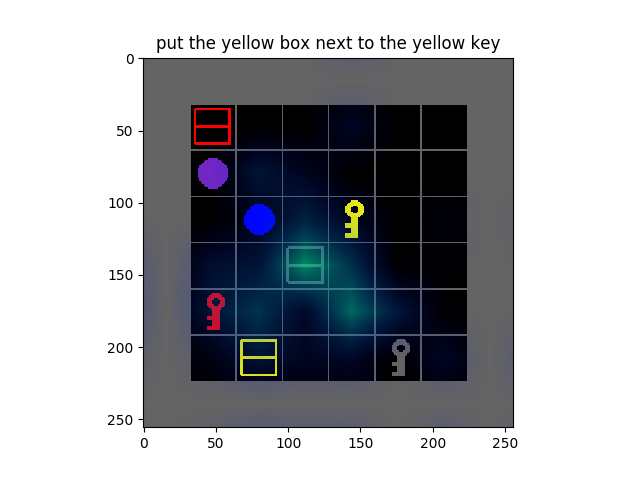}
\end{subfigure}
\begin{subfigure}{0.33\linewidth}
\includegraphics[width=\linewidth]{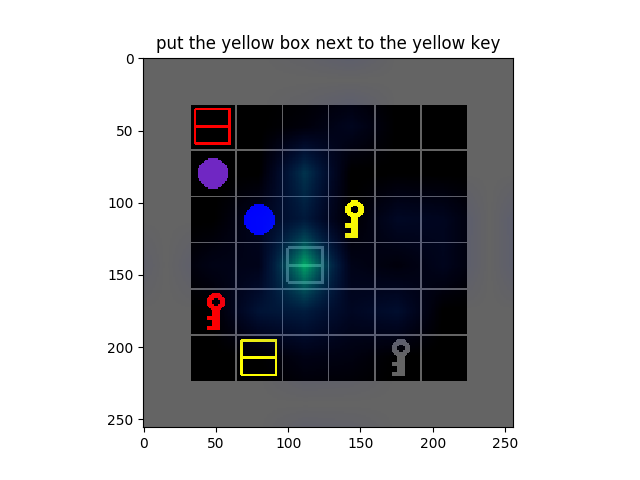}
\end{subfigure}
\caption{Heatmaps for PutNextLocal. They are ordered from left to right and then top to bottom. This shows how the guidance requests evolve over the course of the whole training in one specific example mission.}
\end{figure*}

\end{document}